\documentclass[11pt]{article}
\usepackage[preprint]{acl}

\usepackage{times}
\usepackage{latexsym}

\usepackage[T1]{fontenc}
\usepackage[utf8]{inputenc}

\usepackage{microtype}

\usepackage{inconsolata}

\usepackage{graphicx}

\usepackage{amsmath,amssymb}
\usepackage{amsthm}
\usepackage{algorithm}
\usepackage{algorithmic}
\usepackage{graphicx}
\usepackage{subfig}
\usepackage{multicol} % 导入多列布局宏包
\usepackage{booktabs} % 表格美化
\usepackage{wrapfig}
\usepackage{multirow}
\usepackage{listings}
\usepackage{tcolorbox}
\usepackage{makecell}
\usepackage{enumitem}
\usepackage{stfloats}
\usepackage{float}
\newtheorem{theorem}{Theorem}

\newtheorem{definition}{Definition}
\newtheorem{proposition}{Proposition}

\newtheorem{remark}{Remark}
\newtheorem{corollary}{Corollary}

\usepackage{xcolor}
\usepackage{pifont}
\usepackage{graphicx}
\definecolor{trendred}{RGB}{231, 75, 120}
\definecolor{trendgreen}{RGB}{30, 180, 120}

\newcommand{\up}[1]{%
  \scalebox{0.7}{\textcolor{trendred}{\ding{115}~#1}}%
}
\newcommand{\down}[1]{%
  \scalebox{0.7}{\textcolor{trendgreen}{\ding{116}~#1}}%
}

\usepackage[dvipsnames]{xcolor} % 提供丰富的颜色名称
\usepackage{listings}
\usepackage{multirow}
\usepackage{booktabs}
\definecolor{lightred}{RGB}{255,230,230}

\title{Stable Preference Optimization: A Bilevel Approach to Catastrophic Preference Shift}

\author{
  \textbf{Chengtao Jian}\textsuperscript{1}, \textbf{Kai Yang}\textsuperscript{1}, \textbf{Tianhao Gao}\textsuperscript{1}, \textbf{Wuguang Ni}\textsuperscript{1}, \textbf{Keying Yang}\textsuperscript{1}, \\
  \textbf{Bowen Xiao}\textsuperscript{1}, \textbf{Jiajun Liu}\textsuperscript{1}, \textbf{Ye Ouyang}\textsuperscript{2} \\
  \textsuperscript{1}Tongji University, Shanghai, China \\
  \textsuperscript{2}AsiaInfo Technologies, Beijing, China \\
  \texttt{jct@tongji.edu.cn, kaiyang@tongji.edu.cn}
}

\begin{document}
\maketitle
\begin{abstract}
Direct Preference Learning has emerged as a dominant offline paradigm for preference optimization. Most of these methods are based on the Bradley-Terry (BT) model for pairwise preference ranking, which directly aligns language model with human preference. Prior work has observed a counter-intuitive phenomenon termed likelihood displacement, where the absolute probability of preferred responses decreases simultaneously during training. We demonstrate that such displacement can lead to a more devastating failure mode, which we defined as \textit{Catastrophic Preference Shift}, where the lost preference probability mass inadvertently shifts toward out-of-distribution (OOD) responses. Such a failure mode is a key limitation shared across BT-style direct preference learning methods, due to the fundamental conflict between the unconstrained discriminative alignment and generative foundational capabilities, ultimately leading to severe performance degradation (e.g., SimPO suffers a significant drop in reasoning accuracy from 73.5\% to 37.5\%). We analyze existing BT-style methods from the probability evolution perspective and theoretically prove that these methods exhibit over-reliance on model initialization and can lead to preference shift. To resolve these counter-intuitive behaviors, we propose a theoretically grounded Stable Preference Optimization (SPO) framework that constrains preference learning within a safe alignment region. Empirical evaluations demonstrate that SPO effectively stabilizes and enhances the performance of existing BT-style preference learning methods. SPO provides new insights into the design of preference learning objectives and opens up new avenues towards more reliable and interpretable language model alignment. 

\end{abstract}

\section{Introduction}

Recently, reinforcement learning from human feedback (RLHF)~\citep{ouyang2022training,ziegler2019fine,stiennon2020learning,schulman2017proximal} is a widely used paradigm for preference alignment, it often requires complex reward modeling and substantial computational resources~\citep{ouyang2022training}. To address these challenges, Direct Preference Optimization (DPO)~\citep{rafailov2023direct} and its variants~\citep{wu2025alphadpo,SimPER2025,meng2024simpo} have emerged as a dominant paradigm for aligning models with human preferences, due to its simplicity and effectiveness in leveraging pairwise preference data without the need for explicit reward modeling or reinforcement learning. This approach is based on the Bradley-Terry (BT) model~\citep{bradley1952rank}, which compares the relative preferences between different responses, and has demonstrated significant empirical success in tasks like reasoning~\citep{lai2024step, wang2023making,pang2024iterative} and alignment~\citep{tunstall2023zephyr}.

Despite its widespread adoption and empirical success, DPO exhibits fundamental vulnerabilities that can compromise alignment quality and robustness~\citep{yuan2024advancing,rafailov2024r,tajwar2024preference}. Prior work has shown that the probabilities of preferred and dispreferred responses may decrease simultaneously~\citep{pal2024smaug,rafailov2024r,tajwar2024preference,liu2024provably}, a failure mode known as likelihood displacement~\citep{razin2024unintentional}. Existing work attributes this to factors such as limited model capacity~\citep{tajwar2024preference}, suboptimal initialization of the SFT model~\citep{rafailov2024r}, or input characteristics~\citep{pal2024smaug}. \textit{However, current analyses often rely on specific settings, such as single-token outputs~\citep{razin2024unintentional} or highly similar completions~\citep{pal2024smaug}, making it unclear whether their conclusions generalize to broader settings}. 

A key observation is that DPO is \textit{merely} one special case of a broader family of Bradley-Terry (BT) model-based direct preference learning methods. Recent variants~\citep{wu2025alphadpo,meng2024simpo,xu2024contrastive} introduce modifications to the original DPO objective, yet they remain fundamentally BT-style. This suggests that the limitations observed in DPO~\citep{razin2024unintentional,feng2024towards} may reflect more general properties of these methods rather than issues unique to a particular algorithm. Motivated by this, we first review several recent representative BT model-based direct preference learning methods and show that they can be unified under a single learning objective. We conduct a general theoretical analysis of probability update dynamics for BT model-based preference optimization methods, without making strong assumptions. Finally, we propose a principled bilevel optimization framework that couples supervised fine-tuning and preference optimization through a hierarchical objective, which constrains the preference learning within a safe aligment region. In summary, our contributions are as follows:
\begin{itemize}[left=1pt]
    \item \textbf{Catastrophic Preference Shift.} In stark contrast to prior studies \citep{razin2024unintentional,pal2024smaug} constrained to specific settings, we develop a unified objective framework to analyze the optimization dynamics of BT-style methods. Building on a comprehensive, assumption-light theoretical characterization of their optimization behavior, our work yields a pivotal breakthrough: the identification of a fundamental structural conflict between discriminative alignment and generative foundational capabilities. Most critically, this theoretical lens enables us to pinpoint the general conditions governing a phenomenon termed as \textit{Catastrophic Preference Shift}. Such a discovery fills a critical void in existing literature by providing a principled theoretical explanation for the performance collapse in complex tasks that had previously been documented empirically.
    \item \textbf{Stable Preference Optimization.} Inspired by meta-learning~\citep{franceschi2018bilevel,hospedales2021meta}, we propose a theoretically grounded bilevel optimization method that couples supervised fine-tuning and preference optimization within a single hierarchical objective. In this paradigm, we constrain the preference objective to a safe alignment region, thereby preventing the preference optimization trajectory from drifting beyond the established foundational performance space. Through comprehensive experiments, we demonstrate that SPO substantially improve the performance and stability of existing BT-style methods, especially in complex reasoning tasks.
\end{itemize}

\begin{table*}[t]
\centering
\resizebox{0.99\textwidth}{!}{
\begin{tabular}{lcccccc}
\toprule
\textbf{Method}
& Response
& \(a\)
& \(b\)
& \(\gamma\)
& \(\Omega(\boldsymbol{x},\boldsymbol{y}_w,\boldsymbol{y}_l)\)
\\
\midrule
DPO~\citep{rafailov2023direct}
& \(w / l\)
& \(a_w=a_l=\beta\)
& \(b_w=b_l=\beta\)
& \(0\)
& \(0\)
\\

CPO~\citep{xu2024contrastive}
& \(w / l\)
& \(a_w=a_l=\beta\)
& \(b_w=b_l=0\)
& \(0\)
& \(-\lambda \log \pi_\theta(\boldsymbol{y}_w\mid\boldsymbol{x})\)
\\

R-DPO~\citep{park2024disentangling}
& \(w / l\)
& \(a_w=a_l=\beta\)
& \(b_w=b_l=\beta\)
& \(\alpha(|\boldsymbol{y}_l|-|\boldsymbol{y}_w|)\)
& \(0\)
\\
\multirow{2}{*}{DPOP~\citep{pal2024smaug}}
& $w$
& $a_w=\beta + \lambda \mathcal{I}_w$
& $b_w=\beta + \lambda \mathcal{I}_w$
& $0$
& $0$
\\

& $l$
& $a_l=\beta$
& $b_l=\beta$
& $0$
& $0$
\\

SimPO~\citep{meng2024simpo}
& \(w / l\)
& \(a_w=\frac{\beta}{|\boldsymbol{y}_w|}, a_l=\frac{\beta}{|\boldsymbol{y}_l|}\)
% & \(\frac{\beta}{|\boldsymbol{y}|}\)
& \(b_w=b_l=0\)
& \(\gamma\)
& \(0\)\vspace{3pt}
\\
AlphaDPO~\citep{wu2025alphadpo}
& \(w / l\)
& \(a_w=\frac{\beta}{|\boldsymbol{y}_w|}, a_l=\frac{\beta}{|\boldsymbol{y}_l|}\)
& \(b_w=b_l=0\)
& \(\gamma + \alpha M^\ast(\boldsymbol{x},\boldsymbol{y}_w,\boldsymbol{y}_l)\)
& \(0\)
\\
\bottomrule
\end{tabular}
}
\vspace{-6pt}
\caption{
Instantiation of the unified objective using the score function 
$\mathcal{R}_\theta = a \log \pi_\theta - b \log \pi_{\mathrm{ref}}$. 
Here, $\mathcal{I}_w = \mathcal{I}\!\left[\frac{\pi_{\theta}(\boldsymbol{y}_w \mid \boldsymbol{x})}
{\pi_{\mathrm{ref}}(\boldsymbol{y}_w \mid \boldsymbol{x})} < 1\right]$ 
is an indicator function. \(w\) and \(l\) represent the preferred and dispreferred responses, respectively.
}
\label{tab:instance_unified_objective}
\vspace{-12pt}
\end{table*}

\section{Unified Implicit Reward Framework for Preference Learning}
In this section, we review several recent and representative BT model-based direct preference learning methods for aligning language models with human preferences. We then show that these methods can be unified under a single learning objective.

\subsection{BT-style Direct Preference Learning}

\paragraph{Bradley\-Terry model.}
The Bradley\-Terry (BT) model~\citep{bradley1952rank} is a probabilistic model for pairwise comparisons, where each item $i$ is associated with a positive latent score $p_i > 0$. Given a pair $(i,j)$, the probability that $i$ is preferred over $j$ is defined as
{
\setlength{\abovedisplayskip}{5pt}
\setlength{\belowdisplayskip}{5pt}
\begin{equation}
\Pr(i \succ j)
= \frac{p_i}{p_i + p_j},
\end{equation}
}
where $i \succ j$ denotes that $i$ is preferred over $j$.

\paragraph{Reward Modeling.} Given a preference dataset  $\mathcal{D} = \{(\boldsymbol{x}, \boldsymbol{y}_w, \boldsymbol{y}_l)\}$, where $\boldsymbol{x}$ is the input, $\boldsymbol{y}_w$ is the preferred response, and $\boldsymbol{y}_l$ is the dispreferred response, the reward model $r_\theta$ is trained to assign higher scores to preferred responses. The training objective is defined as
\begin{equation}
\begin{aligned}
% & \mathbb{E}_{(\boldsymbol{x}, \boldsymbol{y}_w, \boldsymbol{y}_l)} \left[ -\log \tfrac{\exp(r_\theta(\boldsymbol{x}, \boldsymbol{y}_w))}{\exp(r_\theta(\boldsymbol{x}, \boldsymbol{y}_w)) + \exp(r_\theta(\boldsymbol{x}, \boldsymbol{y}_l))} \right]
-\log \frac{\exp(r_\theta(\boldsymbol{x}, \boldsymbol{y}_w))}{\exp(r_\theta(\boldsymbol{x}, \boldsymbol{y}_w)) + \exp(r_\theta(\boldsymbol{x}, \boldsymbol{y}_l))}.
\end{aligned}
\end{equation}

\paragraph{DPO.} Direct Preference Optimization~\citep{rafailov2023direct} is a recently proposed method for aligning LLMs with human preferences without the need for explicit reward modeling or reinforcement learning fine-tuning. This approach is a typical instance of the BT model-based direct preference learning methods, which use the BT model to learn from pairwise preference data. ~\citet{rafailov2023direct} identifies the closed-form solution of the RLHF optimization problem, which is defined as
{
\setlength{\abovedisplayskip}{5pt}
\setlength{\belowdisplayskip}{5pt}
\begin{equation}
\begin{aligned}\label{eq:reward_closed_form}
\pi^* (\boldsymbol{y}|\boldsymbol{x}) \propto \pi_{\mathrm{ref}}(\boldsymbol{y}|\boldsymbol{x}) \exp\!\left(\tfrac{1}{\beta} r(\boldsymbol{x}, \boldsymbol{y})\right).
\end{aligned}
\end{equation}
}
By substituting the implicit reward \(r(\boldsymbol{x}, \boldsymbol{y})\) derived from Eq.~\eqref{eq:reward_closed_form} into the BT model, DPO aims to maximize the following objective
{
\setlength{\abovedisplayskip}{5pt}
\setlength{\belowdisplayskip}{5pt}
\begin{equation}
\begin{aligned}
\mathcal{L}_{\mathrm{dpo}}(\theta)
= -\log \sigma \Bigl(
\beta \log \tfrac{\pi_\theta(\boldsymbol{y}_w|\boldsymbol{x})}{\pi_{\mathrm{ref}}(\boldsymbol{y}_w|\boldsymbol{x})}
\\
\qquad\qquad
- \beta \log \tfrac{\pi_\theta(\boldsymbol{y}_l|\boldsymbol{x})}{\pi_{\mathrm{ref}}(\boldsymbol{y}_l|\boldsymbol{x})}
\Bigr),
\end{aligned}
\end{equation}
}
where $\pi_{\mathrm{ref}}$ is a reference policy (often the initial SFT model), $\beta$ is a tunable hyperparameter, and $\sigma(\cdot)$ is the sigmoid function. % that regulates the deviation from the reference model

There are several variants of BT model-based direct preference learning methods that modify the standard DPO method to address limitations or enhance performance. Below, we briefly review several recent representative methods, including CPO~\citep{xu2024contrastive}, R-DPO~\citep{park2024disentangling}, DPOP~\citep{pal2024smaug}, SimPO~\citep{meng2024simpo}, and AlphaDPO~\citep{wu2025alphadpo}, please check Appendix~\ref{app:BT Model-based Direct Preference Learning Methods} for details.

\subsection{A Unified Learning Objective}

To systematically analyze the behavior of these methods, we first formalize the core components of their implicit reward functions.

\begin{definition}[Weighted Log-likelihood Score]
\label{def:weighted_score}
For a given policy $\pi$ and a scaling coefficient $c > 0$, we define the weighted log-likelihood score as
{
\setlength{\abovedisplayskip}{4pt}
\setlength{\belowdisplayskip}{4pt}
\begin{equation}
\phi_{\pi}(\boldsymbol{y} \mid \boldsymbol{x}, c) = c \log \pi(\boldsymbol{y} \mid \boldsymbol{x}).
\end{equation}
}
\end{definition}
Building on this definition, we show that various BT model-based direct preference learning methods can be unified under a single learning objective. Specifically, the implicit reward for a response $\boldsymbol{y}$ can be represented as the discrepancy between the scores of the current and reference policies
{
\setlength{\abovedisplayskip}{4pt}
\setlength{\belowdisplayskip}{4pt}
\begin{equation}
\begin{aligned}
    \mathcal{R}_\theta(\boldsymbol{y} \ |\  \boldsymbol{x}) = \phi_{\pi_\theta}(\boldsymbol{y} \ |\  \boldsymbol{x}, a) - \phi_{\pi_{\mathrm{ref}}}(\boldsymbol{y} \ |\  \boldsymbol{x}, b),
\end{aligned}
\end{equation}
}
where $a$ and $b$ are method-specific coefficients (as detailed in Table~\ref{tab:instance_unified_objective}). For reference-free methods, the objective can be simplified by setting $b=0$. So the unified preference learning objective can be expressed as 
{
\setlength{\abovedisplayskip}{4pt}
\setlength{\belowdisplayskip}{4pt}
\begin{equation}
\setlength{\jot}{0pt}
\begin{aligned}
\label{eq:unified_objective}
&\mathcal{L}_{\mathrm{unified}}(\boldsymbol{\theta}; \mathcal{D})
=-\log \sigma\!\Bigl(
\mathcal{R}_\theta(\boldsymbol{y}_w\mid\boldsymbol{x}) \\
&-\mathcal{R}_\theta(\boldsymbol{y}_l\mid\boldsymbol{x})
-\gamma\Bigr) + \Omega(\boldsymbol{x},\boldsymbol{y}_w,\boldsymbol{y}_l),
\end{aligned}
\end{equation}
}
where $\gamma$ is a margin term, and $\Omega$ is an additional regularization.

\section{Theoretical Analysis}

\subsection{Probability Update Dynamics}

In this subsection, we rigorously analyze the update dynamics of BT model-based preference optimization methods, establishing the theoretical foundation for understanding its optimization behavior.

\begin{theorem}[Probability Update for Preferred and Dispreferred Samples]\label{thm:prob_change}
Let $(\boldsymbol{x}, \boldsymbol{y}_w, \boldsymbol{y}_l)$ be a preference pair. Assuming that the policy $\pi_{\boldsymbol{\theta}}$ is $L$-smooth with respect to $\boldsymbol{\theta}$, the changes in probabilities after a single gradient step of preference optimization with a learning rate $\eta$ can be approximated as
% \begin{subequations}
% \label{eq:delta_pi_dynamics}
% \begin{align}
%     \Delta_w &= \eta (1-\sigma(z)) \pi_{\boldsymbol{\theta}} (\boldsymbol{y}_w|\boldsymbol{x}) \boldsymbol{g}_w^\top (\boldsymbol{g}_w - \boldsymbol{g}_l), \label{eq:delta_pi_y_w} \\
%     \Delta_l &= \eta (1-\sigma(z)) \pi_{\boldsymbol{\theta}} (\boldsymbol{y}_l|\boldsymbol{x}) \boldsymbol{g}_l^\top (\boldsymbol{g}_w - \boldsymbol{g}_l), \label{eq:delta_pi_y_l}
% \end{align}
% \end{subequations}
\begin{align}
    \Delta_w &= \eta (1-\sigma(z)) \pi_{\boldsymbol{\theta}} (\boldsymbol{y}_w|\boldsymbol{x}) \boldsymbol{g}_w^\top (\boldsymbol{g}_w - \boldsymbol{g}_l), \label{eq:delta_pi_y_w} \\
    \Delta_l &= \eta (1-\sigma(z)) \pi_{\boldsymbol{\theta}} (\boldsymbol{y}_l|\boldsymbol{x}) \boldsymbol{g}_l^\top (\boldsymbol{g}_w - \boldsymbol{g}_l). \label{eq:delta_pi_y_l}
\end{align}
where \(\Delta_w \!=\!\Delta \pi_{\boldsymbol{\theta}}(\boldsymbol{y}_w|\boldsymbol{x})\), \(\Delta_l \!=\!\Delta \pi_{\boldsymbol{\theta}}(\boldsymbol{y}_l|\boldsymbol{x})\), \(\boldsymbol{g}_w=a_w\nabla_{\boldsymbol{\theta}} \log \pi_{\boldsymbol{\theta}} (\boldsymbol{y}_w|\boldsymbol{x})\), \(\boldsymbol{g}_l=a_l\nabla_{\boldsymbol{\theta}} \log \pi_{\boldsymbol{\theta}} (\boldsymbol{y}_l|\boldsymbol{x})\), $z = \mathcal{R}_\theta(\boldsymbol{y}_w | \boldsymbol{x})-\mathcal{R}_\theta(\boldsymbol{y}_l | \boldsymbol{x}) -\gamma$. The detailed proof can be found in Appendix~\ref{app:A.1}.
\end{theorem}

\begin{remark}
Theorem~\ref{thm:prob_change} shows that the initial probabilities \(\pi_{\boldsymbol{\theta}}(\boldsymbol{y}_w|\boldsymbol{x})\) and \(\pi_{\boldsymbol{\theta}}(\boldsymbol{y}_l|\boldsymbol{x})\) determine the magnitude of the probability changes for the preferred and dispreferred responses, while the geometric alignment of their log-probability gradients  determines the direction of these changes. Specifically, the direction of \(\Delta_w\) and \(\Delta_l\) depend on the inner product term \(\boldsymbol{g}_w^\top \boldsymbol{g}_l\) (often related to the Neural Tangent Kernel~\citep{jacot2018neural}). When the preferred sample \(\boldsymbol{y}_w\) and the dispreferred sample \(\boldsymbol{y}_l\) exhibit high representational similarity, the term \(\boldsymbol{g}_w^\top (\boldsymbol{g}_w - \boldsymbol{g}_l)\) becomes negative, leading to a decrease in the preferred sample's probability. Combined with Corollary~\ref{corollary:relative_change}, this implies that the dispreferred sample's probability decreases more than the preferred sample's probability, which may render LLMs useless~\citep{duanetal2024negating}. This provides a theoretical explanation for the likelihood displacement phenomenon observed in prior works~\citep{pal2024smaug,razin2024unintentional}.   
\end{remark}

Based on Theorem~\ref{thm:prob_change}, we provide the following two corollaries to further analyze the probability dynamics of preferred and dispreferred samples.

\begin{corollary}[Bound on Preferred and Dispreferred Probability Update]\label{corollary:bound_on_probability_change}
Under the same setting as Theorem~\ref{thm:prob_change}, the change in preferred probability $\Delta \pi_{\boldsymbol{\theta}}(\boldsymbol{y}_w|\boldsymbol{x})$ and dispreferred probability $\Delta \pi_{\boldsymbol{\theta}}(\boldsymbol{y}_l|\boldsymbol{x})$ satisfies
\begin{equation}\label{eq:bound_on_preferred}
\begin{aligned}
\resizebox{0.95\hsize}{!}{$
\Delta_w \geq \eta (1\!-\!\sigma(z)) \pi_{\boldsymbol{\theta}}(\boldsymbol{y}_w|\boldsymbol{x}) \|\boldsymbol{g}_w\|\left(\|\boldsymbol{g}_w\| \! - \! \|\boldsymbol{g}_l\| \right),$}
\end{aligned}
\end{equation}
\begin{equation}
\begin{aligned}\label{eq:bound_on_dispreferred}
\resizebox{0.95\hsize}{!}{$
\Delta_l \leq \eta (1 \!- \!\sigma(z)) \pi_{\boldsymbol{\theta}}(\boldsymbol{y}_l|\boldsymbol{x}) \|\boldsymbol{g}_l\| \left( \|\boldsymbol{g}_w\|\! - \!\|\boldsymbol{g}_l\| \right).$}
\end{aligned}
\end{equation}
% where \(\Delta_w \!=\!\Delta \pi_{\boldsymbol{\theta}}(\boldsymbol{y}_w|\boldsymbol{x})\), \(\Delta_l \!=\!\Delta \pi_{\boldsymbol{\theta}}(\boldsymbol{y}_l|\boldsymbol{x})\), \(\boldsymbol{g}_w=a_w\nabla_{\boldsymbol{\theta}} \log \pi_{\boldsymbol{\theta}} (\boldsymbol{y}_w|\boldsymbol{x})\), \(\boldsymbol{g}_l=a_l\nabla_{\boldsymbol{\theta}} \log \pi_{\boldsymbol{\theta}} (\boldsymbol{y}_l|\boldsymbol{x})\). 
The proof is provided in Appendix~\ref{app:A.2}.
\end{corollary}

\begin{remark}
Corollary~\ref{corollary:bound_on_probability_change} quantifies the conditions for probability changes in both preferred and dispreferred samples. (1) When $\|\boldsymbol{g}_w\| > \|\boldsymbol{g}_l\|$, the lower bound for $\Delta \pi_{\boldsymbol{\theta}}(\boldsymbol{y}_w|\boldsymbol{x})$ is positive, guaranteeing an absolute increase in the preferred response's probability regardless of the gradient alignment. In this case, the upper bound for $\Delta_l$ is also positive, which potentially increases the likelihood of the dispreferred response. (2) Conversely, if $\|\boldsymbol{g}_w\| < \|\boldsymbol{g}_l\|$, the dispreferred probability is guaranteed to decrease, yet the preferred probability $\boldsymbol{y}_w$ may simultaneously decrease if the gradient similarity $\boldsymbol{g}_w^\top \boldsymbol{g}_l$ is sufficiently large. Thus, the BT-based preference learning methods may lead to undesirable outcomes, such as increasing the probability of dispreferred responses or decreasing that of preferred responses.

\end{remark}

To alleviate the risk of decreasing preferred probabilities, according to Eq.~\eqref{eq:bound_on_preferred}, a constraint \(\|\boldsymbol{g}_w\| - \|\boldsymbol{g}_l\|> 0\) should be added to the optimization objective. However, directly constraining the gradients during optimization can be computationally expensive, as it involves calculating Hessian matrices. Instead, we use a surrogate objective to make the optimization more tractable. See Proposition~\ref{prop:surrogate_objective} in Appendix~\ref{app:surrogate} for details.

\begin{corollary}
    [Non-Negativity of Relative Probability Change]\label{corollary:relative_change}
Under the same setting, the difference in probability changes between the preferred and dispreferred samples satisfies
% \begin{equation}
%     \resizebox{0.99\linewidth}{!}{$
% \begin{aligned}
% \Delta \geq& \eta
% \left( 1-\sigma(z) \right) \left\| a_w\sqrt{ \pi_{\boldsymbol{\theta}}(\boldsymbol{y}_w|\boldsymbol{x})} \nabla_{\boldsymbol{\theta}} \log \pi_{\boldsymbol{\theta}}(\boldsymbol{y}_w|\boldsymbol{x}) \right. \\
% & \left. - a_l\sqrt{ \pi_{\boldsymbol{\theta}}(\boldsymbol{y}_l|\boldsymbol{x})} \nabla_{\boldsymbol{\theta}} \log \pi_{\boldsymbol{\theta}}(\boldsymbol{y}_l|\boldsymbol{x}) \right\|^2.
% \end{aligned}
% $}
% \end{equation}
{
\setlength{\jot}{0pt}
\begin{align}
\Delta\! \geq & \eta \left( 1\!-\!\sigma(z) \right) \!\Bigl\| a_w\sqrt{ \pi_{\boldsymbol{\theta}}(\boldsymbol{y}_w|\boldsymbol{x})} \nabla_{\boldsymbol{\theta}}\! \log \pi_{\boldsymbol{\theta}}(\boldsymbol{y}_w|\boldsymbol{x}) \nonumber \\ 
& - a_l\sqrt{ \pi_{\boldsymbol{\theta}}(\boldsymbol{y}_l|\boldsymbol{x})} \nabla_{\boldsymbol{\theta}} \log \pi_{\boldsymbol{\theta}}(\boldsymbol{y}_l|\boldsymbol{x}) \Bigr\|^2, \label{eq:relative_change_full}
\end{align}
}
where \(\Delta=\Delta \pi_{\boldsymbol{\theta}}(\boldsymbol{y}_w|\boldsymbol{x}) - \Delta \pi_{\boldsymbol{\theta}}(\boldsymbol{y}_l|\boldsymbol{x})\). The proof is provided in Appendix~\ref{app:A.3}.
\end{corollary}
\vspace{-4pt}
\begin{remark}
Corollary~\ref{corollary:relative_change} demonstrates that BT-based preference learning methods always guarantees a non-negative relative improvement margin between the preferred sample and the dispreferred one. However, this property does not encourage an absolute increase in the preferred probability, as both probabilities can decrease simultaneously. Such behavior may lead to counter-intuitive training outcomes, where neither the preferred nor the dispreferred sample is truly favored. For instance, if the probability change for the preferred sample is negative, the dispreferred probability decreases even more significantly according to Corollary~\ref{corollary:relative_change}. In this scenario, both probabilities decline as long as the reward margin becomes large. The experical validation of this corollary can be found in Appendix~\ref{Empirical Investigation of Theorem 1}.
\end{remark}

In the next subsection, we will theoretically analyze where the reduced probability mass flows, i.e., which outputs gain probability.

\subsection{Catastrophic Probability Mass Shift}

\begin{theorem}[Probability Mass Shift]\label{thm:mass_shift}
For an input \( \boldsymbol{x} \) with output space \( \mathcal{Y} \), let \( \boldsymbol{y}_w \), \( \boldsymbol{y}_l \), and \( \boldsymbol{y^*} \) be the preferred, dispreferred, and the highest probability outputs, respectively, and define the four disjoint subsets
\(\mathcal{D}_{sw}, \mathcal{D}_{sl}, \mathcal{D}_*, \mathcal{D}_o \subseteq
\mathcal{Y}\) as in Appendix~\ref{app:A.4}. In a single gradient update
\begin{enumerate} \vspace{-4pt}
    \item If \( \Delta \pi_{\boldsymbol{\theta}}(\boldsymbol{y}_w|\boldsymbol{x}) < 0 \), probability mass primarily flows to \( \mathcal{D}_{*} \).\vspace{-4pt}
    \item If \( \Delta \pi_{\boldsymbol{\theta}}(\boldsymbol{y}_l|\boldsymbol{x}) < 0 \), probability mass flows to \( \mathcal{D}_{sw} \) or \( \mathcal{D}_{*} \), depending on their initial probabilities and gradient geometric alignments.\vspace{-4pt}
\end{enumerate}
The detailed proof can be found in Appendix~\ref{app:A.4}.
\end{theorem}

\begin{remark}
Theorem~\ref{thm:mass_shift} clarifies the mechanism of unintentional probability redistribution during BT-based preference learning. It reveals that when the probability of preferred responses declines, or when dispreferred probability decreases without sufficiently favoring the preferred one, the lost probability mass shifts toward OOD generative biases (e.g., those in $\mathcal{D}_{*}$). We define this undesirable misalignment as \textit{Catastrophic Preference Shift}, which represents a severe failure mode that preference probability mass inadvertently shifts toward OOD responses. This phenomenon explains how BT-based preference learning can inadvertently amplify the pre-existing biases or degenerative patterns (such as repetitive text) from the pre-trained or SFT stage, as noted by~\citet{fu2021a,holtzman2019curious}. Such behavior underscores the fragility of unconstrained preference learning in BT-style methods and motivates us to use sophisticated optimization strategies to stabilize the direct preference learning. The empirical validation of this theorem is presented in Appendix~\ref{Empirical Investigation of Theorem 1}.
\end{remark}

The preceding theoretical results characterize the dynamic changes in probabilities, which reveals several critical vulnerabilities in BT-based preference learning: (1) \textit{Initialization Bias}: They are over-reliance on the model bias, which can propagate through the gradient field and lead to suboptimal convergence; (2) \textit{Likelihood Displacement}: They only guarantee a relative probability improvement of preferred over dispreferred responses, meaning that both the preferred and dispreferred sample's probability may decrease simultaneously; (3) \textit{Unintended Preference Shift}: Such absolute declines inevitably cause preference shift toward spurious or non-preferred outputs, potentially amplifying pre-existing biases or degenerative patterns. These vulnerabilities highlight the need for more robust optimization techniques that can effectively balance preference alignment with the preservation of foundational generative capabilities. In the next section, we introduce Stable Preference Optimization (SPO), a bilevel optimization framework designed to address these challenges.

\section{Stable Preference Optimization}

\subsection{Motivation and Problem Formulation}

Our theoretical analysis reveals that standard BT-based preference optimization, primarily focuses on the relative margin between preferred and dispreferred responses. However, as derived in Theorem \ref{thm:prob_change} and Corollary~\ref{corollary:relative_change}, this discriminative objective creates a fundamental conflict with the LLM's generative objective in the gradient space. When the gradients of preferred and dispreferred samples are highly representational similar, the alignment pressure inevitably drags down the absolute likelihood of responses. According to Theorem \ref{thm:prob_change} and Theorem \ref{thm:mass_shift}, this issue is particularly dangerous when the initial probability of the preferred response is low. In this case, the probability mass lost from the non-preferred samples is redistributed to responses in non-preferred regions with higher initial density, rather than to the preferred response. In an extreme scenario, this may cause the model to drift away from the foundational generative space (the capabilities established during SFT). 

Such a conflict suggests that preference alignment should not be treated as a unconstrained single-level task. To prevent the catastrophic preference shift issue described by Theorem \ref{thm:mass_shift}, alignment should be formulated as a search for an optimal model within a safe region where the preferred response is anchored as the high-density mode. This naturally leads to a hierarchical decision-making structure. Mathematically, this structure can be elegantly captured by bilevel optimization~\citep{jiao2023asynchronous}. Unlike joint optimization, the bilevel framework allows the upper-level alignment objective to receive feedback from the lower-level problem, effectively thereby preventing the optimization process from compromising foundational language modeling performance.

Consequently, we can formulate SPO as a bilevel problem. Specifically, the primary (upper-level) objective is to search an optimal model via preference objectives like DPO, while the lower-level objective serves as an constraint to preserves the foundational generative capabilities established during SFT. To further stabilize the optimization, we also incorporate the surrogate regularization term from Proposition~\ref{prop:surrogate_objective} into the upper-level objective. The final bilevel formulation is given by
{
\setlength{\abovedisplayskip}{5pt}
\setlength{\belowdisplayskip}{5pt}
\begin{equation}
    \resizebox{0.99\linewidth}{!}{$
\setlength{\jot}{0pt}
\begin{aligned}
& \min_{\boldsymbol{\theta}}\  \mathcal{L_{\mathrm{pref}}} := 
    \mathbb{E}_{(\boldsymbol{x}, \boldsymbol{y}_w, \boldsymbol{y}_l) \sim \mathcal{D}_{\mathrm{tr}}} 
    \mathcal{L}_{\mathrm{unified}}(\boldsymbol{x}, \boldsymbol{y}_w, \boldsymbol{y}_l; \boldsymbol{\theta}) \\
     & \qquad \qquad + \gamma \underbrace{
        \mathbb{E}_{(\boldsymbol{x}, \boldsymbol{y}_w, \boldsymbol{y}_l) \sim \mathcal{D}_{\mathrm{tr}}} 
        \mathcal{L}_{\text{reg}}(\boldsymbol{x}, \boldsymbol{y}_w; \boldsymbol{\theta})
    }_{\text{Theorem \ref{thm:prob_change} \& Proposition~\ref{prop:surrogate_objective}}} \\
& \text{s.t.} \  
\boldsymbol \theta \in \arg\min_{\boldsymbol \theta^{\prime}}
    \underbrace{
        \mathbb{E}_{(\boldsymbol{x}, \boldsymbol{y}_w) \sim \mathcal{D}_{\mathrm{sft}}}
         \mathcal{L}_{\mathrm{sft}}(\boldsymbol{x}, \boldsymbol{y}_w;\boldsymbol{\theta}^{\prime}) 
    }_{\text{Theorem~\ref{thm:mass_shift}}}.
\end{aligned}
$}
\label{eq:bilevel}
\end{equation}
}
In this formulation, the upper-level objective steers the model towards human preferences within the safe alignment region (feasible set) defined by the lower-level constraint. This structure embodies both \textit{cooperative} and \textit{competitive} dynamics: the lower-level generative objective cooperatively preserves a strong language modeling capability to support the upper-level preference objective, while the upper-level objective competitively discriminates between preferred and dispreferred responses to achieve optimal alignment. Next, we develop an efficient penalty-based algorithm to facilitate the practical application of SPO in LLM fine-tuning.

\subsection{The Bilevel Learning Algorithm}

\begin{figure}[t]
\centering
\includegraphics[width=0.48\textwidth]{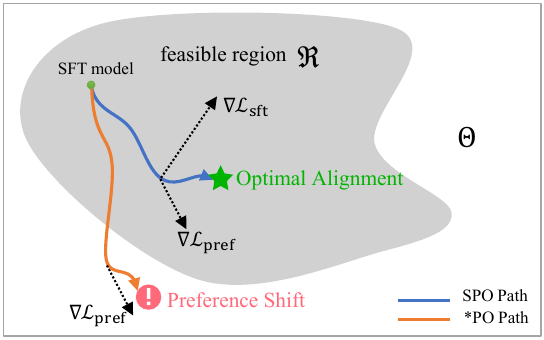}
\caption{Geometric illustration of SPO. Standard preference optimization paths (orange) often drift out of the safe alignment region $\mathfrak{R}$ (gray), leading to preference shift. In contrast, SPO (blue) utilizes a bilevel structure constrain updates within $\mathfrak{R}$, preserving foundational generative capabilities while aligning with preferences.} 
\vspace{-2pt}
\label{fig:grad_analysis}
\end{figure}

Inspired by previous penalty-based bilevel optimization algorithms~\citep{shen2023penalty, liu2022bome}, we reformulate the bilevel problem with penalty functions via lower-level value-function-based constraints. Specifically, following~\citet{nouiehed2019solving, liu2022bome,kwon2023fully}, we replace the lower-level optimization problem with a constraint, leveraging the optimal lower-level value function \(\mathcal{L}_{\mathrm{sft}}^{*}(\boldsymbol{x}, \boldsymbol{y}_w) = \min_{\boldsymbol{\theta}^{\prime}} \mathcal{L}_{\mathrm{sft}}(\boldsymbol{x}, \boldsymbol{y}_w; \boldsymbol{\theta}^{\prime})\). This results in the following relaxed single-level constrained optimization problem 
{
\setlength{\abovedisplayskip}{6pt}
\setlength{\belowdisplayskip}{6pt}
\begin{equation} \label{eq:constrained_opt_refined} 
\min_{\boldsymbol{\theta}} \mathcal{L}_{\mathrm{pref}}(\boldsymbol{x}, \boldsymbol{y}_w, \boldsymbol{y}_l; \boldsymbol{\theta})\ \text{s.t.}\ \boldsymbol{\theta} \in \mathfrak{R}(\boldsymbol{x}, \boldsymbol{y}_w),
\end{equation}
}
where $\mathfrak{R}(\boldsymbol{x}, \boldsymbol{y}_w) = \{ \boldsymbol{\theta} \mid \mathcal{L}_{\mathrm{sft}}(\boldsymbol{x}, \boldsymbol{y}_w; \boldsymbol{\theta}) \leq \mathcal{L}_{\mathrm{sft}}^{*} (\boldsymbol{x}, \boldsymbol{y}_w) + \epsilon \}$ denotes the feasible region. As illustrated in Figure~\ref{fig:grad_analysis}, this region constrains the parameters $\boldsymbol{\theta}$ update within a safe aligment region, addressing the likelihood displacement issue.

To solve this constrained problem, we further penalize the constraint onto the upper-level optimization problem. Following ~\citet{pan2024scalebio,shen2023penalty}, we introduce a penalty term with coefficient \(\lambda > 0\), resulting in
\begin{equation} \label{eq:penalized_opt}
    \resizebox{0.87\linewidth}{!}{$
    \setlength{\jot}{0pt}
\begin{aligned}
\min_{\boldsymbol{\theta}} &\ \mathcal{L}_{\mathrm{spo}}(\boldsymbol{x}, \boldsymbol{y}_w, \boldsymbol{y}_l; \boldsymbol{\theta}) := \mathcal{L}_{\mathrm{pref}}(\boldsymbol{x}, \boldsymbol{y}_w, \boldsymbol{y}_l; \boldsymbol{\theta}) \\
& + \lambda \left( \mathcal{L}_{\mathrm{sft}}(\boldsymbol{x}, \boldsymbol{y}_w; \boldsymbol{\theta}) - \mathcal{L}_{\mathrm{sft}}^{*}(\boldsymbol{x}, \boldsymbol{y}_w) \right),
\end{aligned}
$}
\end{equation}
where \(\mathcal{L}_{\mathrm{pref}}(\boldsymbol{x}, \boldsymbol{y}_w, \boldsymbol{y}_l; \boldsymbol{\theta})\) includes both the preference objective and the regularization term as defined in Eq.~\eqref{eq:bilevel}. This penalized objective balances the preference optimization with the lower-level constraint, which mitigates the over-reliance on model initialization. The optimization problem in Eq.~\eqref{eq:penalized_opt} can be solved using gradient-based methods. While the penalty term involves the optimal lower-level value function \(\mathcal{L}_{\mathrm{sft}}^{*}(\boldsymbol{x}, \boldsymbol{y}_w)\), which requires solving a lower-level minimization problem, its gradient with respect to variable \(\boldsymbol{\theta}\) is zero, i.e., \(\nabla_{\boldsymbol \theta} \mathcal{L}_{\mathrm{sft}}^{*}(\boldsymbol{x}, \boldsymbol{y}_w) = 0\). This is because \(\mathcal{L}_{\mathrm{sft}}^{*}(\boldsymbol{x}, \boldsymbol{y}_w)\) is independent of \(\boldsymbol{\theta}\) and only depends on the optimal solution \(\boldsymbol{\theta}^{*}\) of the lower-level problem. 

Based on this, given the number of outer iterations \(t\), we approximate the lower-level optimization problem using K-step gradient descent:
{
\setlength{\abovedisplayskip}{5pt}
\setlength{\belowdisplayskip}{5pt}
\begin{equation} \label{eq:k_step_gd}
\boldsymbol{\theta}^{\prime(k+1)}_t = \boldsymbol{\theta}^{\prime(k)}_t - \eta_{\theta^{\prime}} \nabla_{\boldsymbol{\theta}^{\prime}} \mathcal{L}_{\mathrm{sft}}(\boldsymbol{x}, \boldsymbol{y}_w ; \boldsymbol{\theta}^{\prime(k)}_t),
\end{equation}
}
where \(\eta_{\theta^{\prime}}\) is the learning rate for the lower-level updates. After \(K\) iterations, \(\boldsymbol{\theta}^{\prime(K)}_t\) approximates \(\boldsymbol{\theta}^{*}\), and we estimate \(\mathcal{L}_{\mathrm{sft}}^{*}(\boldsymbol{x}, \boldsymbol{y}_w) \approx \mathcal{L}_{\mathrm{sft}}(\boldsymbol{x}, \boldsymbol{y}_w; \boldsymbol{\theta}^{\prime(K-1)}_t)\) for each given \(\boldsymbol{\theta}_t\). Then we can solve Eq.~\eqref{eq:penalized_opt} via gradient descent, and the parameter update is as follows
{
\setlength{\abovedisplayskip}{5pt}
\setlength{\belowdisplayskip}{5pt}
\begin{equation} \label{eq:variable_updates}
\boldsymbol{\theta}_{t+1} = \boldsymbol{\theta}_{t} - \eta_{\theta} \nabla_{\boldsymbol{\theta}} \mathcal{L}_{\mathrm{spo}}(\boldsymbol{x}, \boldsymbol{y}_w, \boldsymbol{y}_l; \boldsymbol{\theta}_{t}),
\end{equation}
}
where \(\eta_{\theta}\) is the learning rate for \(\boldsymbol{\theta}\). The overall algorithm is summarized in Algorithm~\ref{alg:bilevel_dpo}. In practice, to reduce computational overhead,
we use a more efficient algorithm (see Algorithm~\ref{alg:eff_bilevel_dpo}) that avoids the inner loop for lower-level optimization. 

\begin{figure*}[!t]
\centering
\vspace{-12pt}
\begin{equation} 
    \resizebox{0.88\linewidth}{!}{$
\begin{aligned} \label{eq:grad_theta_doc}
\nabla_{\boldsymbol{\theta}} \mathcal{L}_{\mathrm{spo}} &= -(1 - \sigma(z)) \underbrace{\left( a_w\nabla_{\boldsymbol{\theta}} \log \pi_{\boldsymbol{\theta}}(\boldsymbol{y}_w|\boldsymbol{x}) - a_l\nabla_{\boldsymbol{\theta}} \log \pi_{\boldsymbol{\theta}}(\boldsymbol{y}_l|\boldsymbol{x}) \right)}_{\text{(a) preference discrimination} } + \gamma\underbrace{\nabla_{\boldsymbol \theta}\mathcal{L}_{\text{reg}}(\boldsymbol{x}, \boldsymbol{y}_w; \boldsymbol{\theta})}_{\text{(b) gradient regularization}}\\ 
&\quad - \lambda \underbrace{\left( \nabla_{\boldsymbol{\theta}} \log \pi_{\boldsymbol{\theta}}(\boldsymbol{y}_w|\boldsymbol{x}) - \nabla_{\boldsymbol{\theta}} \log \pi_{\boldsymbol{\theta}^{*}}(\boldsymbol{y}_w|\boldsymbol{x}) \right)}_{\text{(c) feasible region projection}},
\end{aligned} 
$}
\end{equation}\vspace{-12pt}
\end{figure*}

\begin{algorithm}[t]
\caption{Stable Preference Optimization}\label{alg:bilevel_dpo}
\begin{algorithmic}[1]
\STATE \textbf{Input}: Preference dataset \(\mathcal{D}_{\mathrm{tr}}\), SFT dataset \(\mathcal{D}_{\mathrm{sft}}\), initial parameters \(\boldsymbol{\theta}_0\), penalty coefficient \(\lambda\), learning rates \(\eta_{\theta}\), \(\eta_{\theta '}\), number of outer iterations \(T\), lower-level update step \(K\).
\STATE Initialize \(\boldsymbol{\theta}_0 \leftarrow \boldsymbol{\theta}_0\).
\FOR{\(t = 0, 1, \dots, T-1\)}
    \STATE Initialize \(\boldsymbol{\theta}^{\prime(0)}_t \leftarrow \boldsymbol{\theta}_t\).
    \FOR{\(k = 0, 1, \dots, K-1\)}
        \STATE Sample batch \((\boldsymbol{x}, \boldsymbol{y}_w) \sim \mathcal{D}_{\mathrm{sft}}\).
        \STATE Update \(\boldsymbol{\theta}^{\prime(k+1)}_t\) via Eq.~\eqref{eq:k_step_gd}.
    \ENDFOR
    \STATE Set \(\boldsymbol{\theta}^* = \boldsymbol{\theta}^{\prime(K-1)}_t\).
    \STATE Compute \(\mathcal{L}_{\mathrm{sft}}^{*}(\boldsymbol{x}, \boldsymbol{y}_w) \approx \mathcal{L}_{\mathrm{sft}}(\boldsymbol{x}, \boldsymbol{y}_w; \boldsymbol{\theta}^{*})\).
    \STATE Sample mini-batch \((\boldsymbol{x}, \boldsymbol{y}_w, \boldsymbol{y}_l)\).
    \STATE Update \(\boldsymbol{\theta}_{t+1}\) via Eq.~\eqref{eq:variable_updates}.
\ENDFOR
\STATE \textbf{Output}: Parameters \(\boldsymbol{\theta}_{T-1}\).
\end{algorithmic}
\end{algorithm}

\subsection{Gradient Analysis} \label{sec:Analysis}
To understand how SPO mitigates the limitations, we analyze the gradients of the optimization objective in SPO. Given the optimization objective in Eq.~\eqref{eq:penalized_opt}, the gradient with respect to \(\boldsymbol{\theta}\) is present Eq.~\eqref{eq:grad_theta_doc}. Derivation can be found in Appendix~\ref{app:A.5}

\textbf{What does SPO update do?} The SPO update, as defined by the gradient in Eq.~\eqref{eq:grad_theta_doc}, addresses potential limitations of BT model-based preference learning identified in Theorems \ref{thm:prob_change} and \ref{thm:mass_shift}. The gradient terms (a)--(c) in Eq.~\eqref{eq:grad_theta_doc} work as follows:\vspace{-2pt}
\begin{itemize}[left=1pt]
    \item (a) Preference Discrimination: This term promotes the model to favor preferred responses over dispreferred ones by enlarging the relative log-probability margin between the preferred response and the dispreferred response. \vspace{-2pt}
    \item (b) Gradient Regularization: This term regularizes the gradient norms to stabilize the optimization process. It acts as a safeguard against undesirable probability mass shift toward out-of-distribution responses, such as non-preferred outputs, as highlighted in Theorem~\ref{thm:mass_shift}.\vspace{-2pt}
    \item (c) Feasible Region Projection: This is the core structural constraint derived from the bilevel formulation. As visualized in Figure~\ref{fig:grad_analysis}, it constrains the update direction to remain within the safe alignment region, mitigating the preference shift.\vspace{-2pt}
\end{itemize}

In summary, while standard methods are often diverged from the optimal generative manifold due to the unconstrained preference optimization, SPO maintains optimization stability. The interplay between these three terms ensures that the model effectively navigates the preference lanscape while preserving its foundational generative capabilities.

\begin{table*}[!t]
\centering
\small
\caption{Main results on AlpacaEval 2 and GSM8K across different base models.}
\vspace{-6pt}
\label{tab:main_results}
\resizebox{0.99\textwidth}{!}{
\begin{tabular}{l ccc ccc ccc}
\toprule
\multirow{3}{*}{\textbf{Method}}
& \multicolumn{3}{c}{\textbf{Qwen-2.5-0.5B}}
& \multicolumn{3}{c}{\textbf{LLaMA-3.2-1B}}
& \multicolumn{3}{c}{\textbf{LLaMA-3-8B}} \\
\cmidrule(lr){2-4}\cmidrule(lr){5-7}\cmidrule(lr){8-10}

& \multicolumn{2}{c}{\textbf{AlpacaEval 2}} & \textbf{GSM8K}
& \multicolumn{2}{c}{\textbf{AlpacaEval 2}} & \textbf{GSM8K}
& \multicolumn{2}{c}{\textbf{AlpacaEval 2}} & \textbf{GSM8K} \\
\cmidrule(lr){2-3}\cmidrule(lr){4-4}
\cmidrule(lr){5-6}\cmidrule(lr){7-7}
\cmidrule(lr){8-9}\cmidrule(lr){10-10}

& \textbf{WR (\%)} & \textbf{LC (\%)} & \textbf{Acc. (\%)}
& \textbf{WR (\%)} & \textbf{LC (\%)} & \textbf{Acc. (\%)}
& \textbf{WR (\%)} & \textbf{LC (\%)} & \textbf{Acc. (\%)} \\
\midrule

SFT 
& 2.0 & 2.1 & 42.7
& 3.5 & 3.2 & 49.8
& 11.9 & 11.2 & 73.5 \\
\midrule

DPO  
& 4.2 & 4.6 & 35.2
& 3.9 & 4.6 & 49.6
& 15.4 & 14.6  & 63.8 \\
DPO+SPO  
& 4.4 \up{0.2} & 4.8 \up{0.2} & 43.2 \up{8.0}
& 4.2 \up{0.3} & 4.9 \up{0.3} & 54.4 \up{4.8}
& 16.0 \up{0.6} & 17.4 \up{2.8}  & 76.5 \up{12.7} \\
\midrule

CPO  
& 2.3 & 3.5 & 33.5
& 4.6 & 3.7 & 48.4
& 16.9 & 16.4 & 73.2 \\
CPO+SPO  
& 3.2 \up{0.9} & 3.7 \up{0.2} & 41.3 \up{7.8}
& 4.6 \up{0.0} & 3.9 \up{0.2} & 49.9 \up{1.5}
& 16.7 \down{0.2} & 17.1 \up{0.7} & 77.1 \up{3.9} \\
\midrule

DPOP  
& 3.7 & 4.2 & 43.1
& 3.4 & 2.7 & 53.6
& 13.2 & 12.7 & 74.6 \\
DPOP+SPO  
& 3.6 \down{0.1} & 4.4 \up{0.2} & 44.1 \up{1.0}
& 3.6 \up{0.2} & 3.2 \up{0.5} & 53.9 \up{0.3}
& 13.0 \down{0.2} & 13.3 \up{0.6} & 77.2 \up{2.6} \\
\midrule

RDPO
& 3.2 & 3.8 & 34.3
& 3.6 & 2.8 & 42.2
& 14.7 & 13.7 & 65.4 \\
RDPO+SPO  
& 3.3 \up{0.1} & 3.9 \up{0.1} & 43.8 \up{9.5}
& 4.0 \up{0.4} & 3.5 \up{0.7} & 55.0 \up{12.8}
& 15.7 \up{1.0} & 17.4 \up{3.7} & 75.5 \up{10.1} \\
\midrule

SimPO  
& 5.5 & 4.1 & 22.3
& 6.4 & 6.5 & 26.9
& 19.6 & 21.8 & 37.4 \\
SimPO+SPO  
& 6.5 \up{1.0} & 5.6 \up{1.5} & 43.9 \up{21.6}
& 8.2 \up{1.8} & 7.5 \up{1.0} & 52.2 \up{25.3}
& 22.6 \up{3.0} & 26.5 \up{4.7} & 76.0 \up{38.6} \\
\midrule

AlphaDPO  
& 4.5 & 4.2 & 34.7
& 8.2 & 8.6 & 32.9
& 20.4 & 22.3 & 33.5 \\
AlphaDPO+SPO  
& 6.7 \up{2.5} & 6.0 \up{1.8} & 40.3 \up{5.6}
& 9.0 \up{0.8} & 8.8 \up{0.2} & 42.3 \up{9.4}
& 22.0 \up{1.6} & 26.6 \up{4.3} & 70.2 \up{36.7} \\

\bottomrule
\end{tabular}
}
\vspace{-6pt}
\end{table*}

\section{Experiments} \label{sec:experiment}

\subsection{Experimental Setup}\label{sec:exp-setup}

\paragraph{Models and Training Datasets.} We evaluate our approach using three language models: Qwen-2.5-0.5B~\citep{qwen2_5}, LLaMA-3.2-1B~\citep{grattafiori2024llama}, and LLaMA-3-8B~\citep{grattafiori2024llama}. For instruction-following tasks, we utilize the UltraFeedback dataset~\citep{cui2023ultrafeedback}, where all models are first fine-tuned on its SFT split to obtain base SFT checkpoints, followed by preference optimization on the pairwise preference split. For reasoning tasks, we leverage the questions from the GSM8K training set~\citep{cobbe2021training} and generate high-quality reasoning chains using DeepSeek-R1-Distill-Qwen-7B~\citep{guo2025deepseek} to construct the SFT and preference data. The details of preference dataset construction are provided in Appendix~\ref{app:experimental_details}.
\vspace{-4pt}
\paragraph{Evaluation Benchmarks.} 
We evaluate our models using two widely adopted benchmarks: AlpacaEval 2.0~\citep{dubois2024length} and GSM8K~\citep{cobbe2021training}. The AlpacaEval benchmark measure instruction-following performance, while GSM8K benchmark measure reasoning ability. For AlpacaEval, we report both the raw win rate and the length-controlled win rate. For GSM8K, we report the accuracy. Detailed evaluation protocols are provided in Appendix~\ref{app:experimental_details}.

\vspace{-4pt}
\paragraph{Baselines and SPO Integration.}
Our goal is to evaluate whether SPO can improve the stability and effectiveness of existing BT-based direct preference optimization methods.
We therefore apply SPO as a plug-in on top of several representative methods: DPO~\citep{rafailov2023direct}, CPO~\citep{xu2024contrastive}, RDPO~\citep{park2024disentangling}, and DPOP~\citep{pal2024smaug}, SimPO~\citep{meng2024simpo}, AlphaDPO~\citep{wu2025alphadpo}. For each method, we compare its vanilla version with SPO-enhanced version. 

\begin{figure*}[t]
    \centering
    \includegraphics[width=0.99\textwidth]{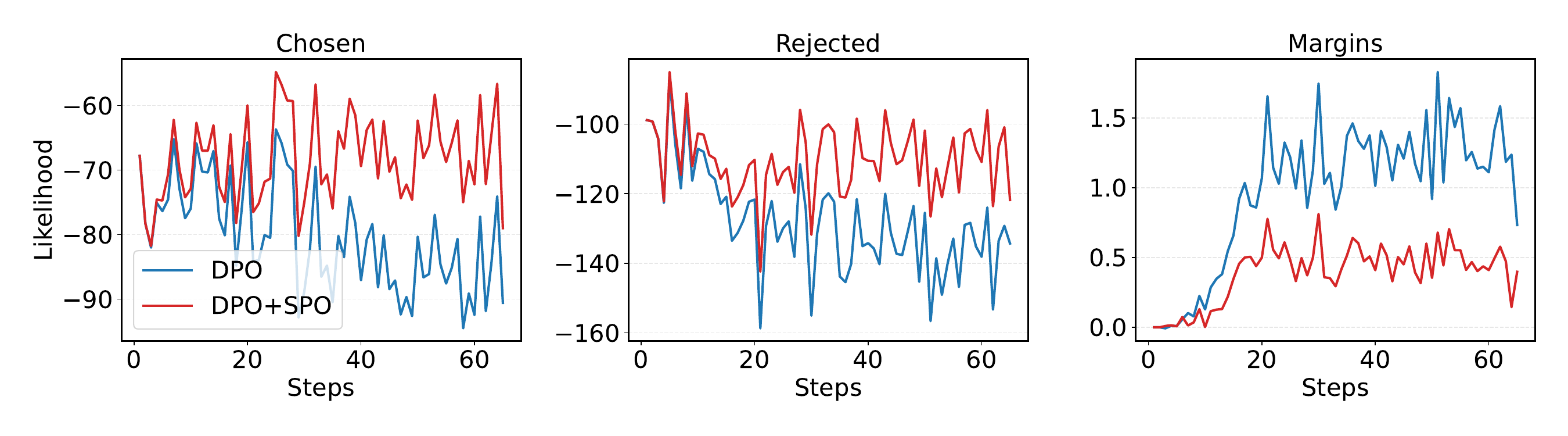}
    \vspace{-15pt}
    \caption{Training dynamics of DPO and DPO+SPO.}
    \label{fig:dpo_spo_training_dynamics}
    \vspace{-12pt}
\end{figure*}

\subsection{Main Results}

\paragraph{SPO consistently enhanced existing BT-style preference optimization methods.} As shown in Table~\ref{tab:main_results}, SPO improves the performance of existing BT-style methods in almost all settings. While all baselines benefit from preference optimization, SPO-enhanced methods achieve improved performance by stabilizing probability updates. These results further validate our theoretical findings and demonstrate the effectiveness of SPO. Specifically, SPO improves the best baseline by an average 2.1\% in length-controlled win rate and 1.6\% in raw win rate. Moreover, on challenging reasoning tasks, SPO improves the best baseline by average 1.3\% accuracy. Moreover, we observe that SimPO and AlphaDPO drop significantly in GSM8K accuracy without SPO, indicating catastrophic preference shift issues. Integrating SPO substantially mitigates this issue, leading to remarkable improvements. We also provide case studies to illustrate this catastrophic preference shift observed in reasoning tasks.Please refer to Appendix~\ref{app:case_study}.
\vspace{-4pt}
\paragraph{SPO stabilizes existing BT-style preference optimization methods.} As shown in Figure~\ref{fig:dpo_spo_training_dynamics}, we can observe that DPO+SPO maintains higher and more stable probabilities for preferred responses compared to vanilla DPO, while the vanilla DPO exhibits significant drops in preferred and dispreferred probabilities during training. Although vanilla DPO successfully improves the relative reward margin between preferred and dispreferred responses, it fails to enhance the reasoning accuracy, which drops from 73.5\% to 63.8\% in the LLaMA3-8B model. Instead, DPO+SPO maintains stable optimization dynamics, leading to a substantial improvement in GSM8K accuracy to 76.5\%. Similar trends are observed for other BT-style methods (see Appendix~\ref{app:additional_experiments}). Such a performance-reward gap indicates unstable optimization dynamics, which can be explained by our theoretical analysis in Theorem~\ref{thm:prob_change} and Corollary~\ref{corollary:relative_change}. Consequently, the policy model optimized by vanilla BT-style methods deviates from the safe alignment region, ultimately leading to preference shift. 
\paragraph{Additional Experiments.} We also evaluate the downstream tasks on the HuggingFace Leadboards and observe similar improvements trends when integrating SPO with existing BT-style methods (see Appendix~\ref{app:evaluation_on_downstream_tasks}). Moreover, we extend SPO to SFT-free setting (see Appendix~\ref{app:extend_to_sft_free}), and we observe that SPO represents a promising direction in SFT-free preference optimization. These additional experiments further validate the effectiveness and versatility of SPO.

\subsection{Ablation Analysis}
\begin{figure*}[ht]
\centering 
\subfloat{\includegraphics[width=0.33\textwidth]{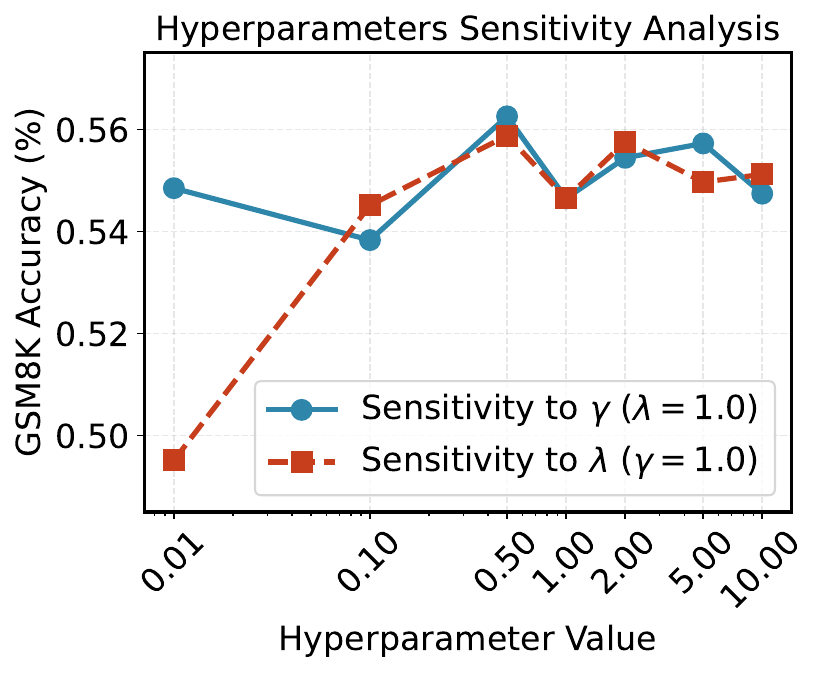}}\hfill
\subfloat{\includegraphics[width=0.33\textwidth]{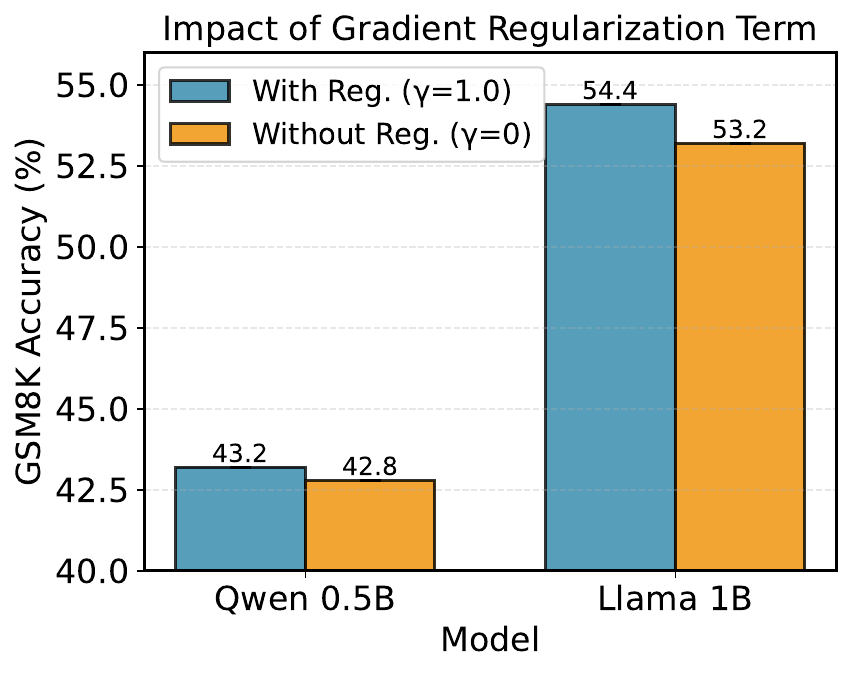}}\hfill
\subfloat{\includegraphics[width=0.33\textwidth]{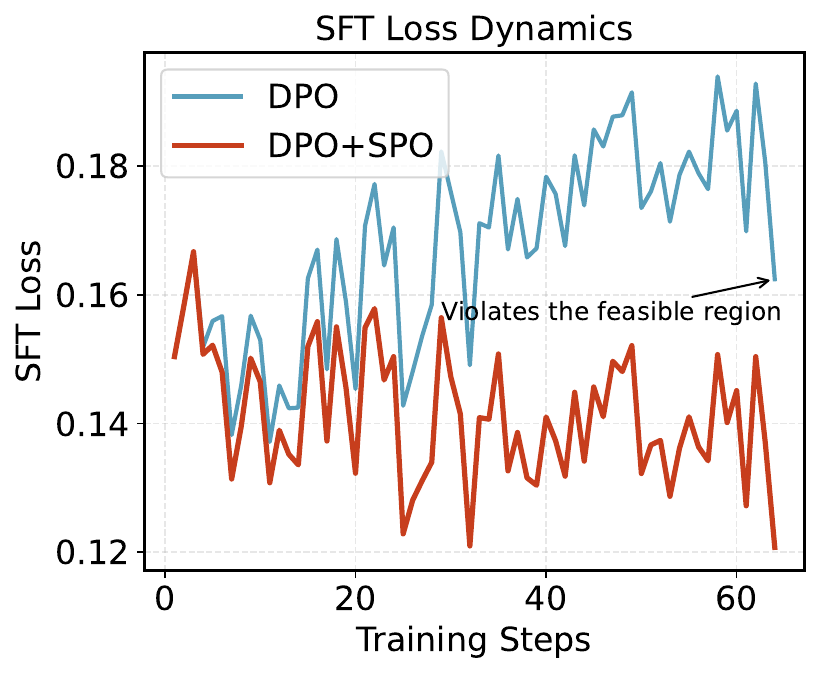}}
\vspace{-6pt}
\caption{Sensitivity to \(\lambda\) and \(\gamma\) (left), effect of gradient regularization (middle) and SFT loss dynamics (right).}
\label{fig:effect_hp}
\vspace{-12pt}
\end{figure*}

To better understand the impact of SPO's hyperparameters, we conduct ablation studies on GSM8K using the Qwen2.5-0.5B and LLaMA-3.2-1B models. Since all these methods are based on the BT model, we select DPO as the representative method for ablation studies. 
\vspace{-4pt}
\paragraph{Effect of hyperparameter \(\lambda\) and \(\gamma\).} As illustrated in Figure~\ref{fig:effect_hp} (left), we can observe that SPO exhibits more sensitivity to the penalty coefficient $\lambda$ than to the regularization weight $\gamma$. Performance significantly degrades when $\lambda$ is too small, as the unconstrained preference learning tends to increase the reward margin by simultaneously reducing the absolute probabilities of both responses. However, SPO is relatively robust to variations in \(\gamma\), this is mainly because the lower-level constraint of our bilevel learning algorithm is primarily enforced through the penalty coefficient \(\lambda\), rather than being directly influenced by \(\gamma\). Overall, the empirical observations are consistent with and further support the theoretical insights of SPO.
\vspace{-4pt}
\paragraph{Impact of Gradient Regularization.} As shown in Figure~\ref{fig:effect_hp} (middle), we compare the performance of DPO+SPO with and without gradient regularization on GSM8K, which indicates that incorporating gradient regularization leads to a small but consistent improvement in accuracy. This suggests that gradient regularization can further enhance the overall performance.
\vspace{-4pt}
\paragraph{Feasible Region Preservation.} We compare the SFT loss trajectories of vanilla DPO and DPO+SPO in Figure~\ref{fig:effect_hp} (right). We can observe that SPO maintains a stable SFT loss throughout the training process, indicating that SPO does not violate the feasible region $\mathfrak{R}$. In contrast, vanilla DPO exhibits a persistent upward trend, suggesting that it mathematically deviates from the safe region, which aligns with our theoretical analysis.

\section{Related Work}

\vspace{-4pt}
\paragraph{Likelihood Displacement in DPO.} Recent works~\citet{ren2025learning_dynamics_LLM,pal2024smaug} demonstrate that DPO may unintentionally reduce the probability of preferred responses, a phenomenon known as likelihood displacement~\citep{razin2024unintentional}, especially when completions are semantically similar~\citep{cho2025rethinking}.
\citet{feng2024towards} analyze the gradient field of DPO, highlighting its tendency to prioritize decreasing the likelihood of dispreferred response. While several DPO variants~\citep{yang2025dposhiftshiftingdistributiondirect,pal2024smaug} have been proposed to mitigate this issue, a comprehensive theoretical analysis of probability evolution in a more general remains lacking. \textit{Our work addresses this gap by providing a rigorous theoretical investigation on a broader class of BT model-based preference optimization methods.}

\vspace{-4pt}
\paragraph{Bilevel Optimization.}
Bilevel optimization provides a principled framework for capturing hierarchical decision-making processes~\citep{liu2025argus, jiao2023asynchronous}, where one optimization process is nested within the constraint set of another. This paradigm has recently emerged as a versatile tool for various complex settings, including robust optimization~\citep{yang2008distributed,qian2019robust}, meta-learning~\citep{franceschi2018bilevel,hospedales2021meta,gu2021nonconvex}, and Stackelberg games~\citep{stackelbergGame}. In the context of large language models, bilevel and multi-level structures have demonstrated significant promise in areas such as security alignment~\citep{jiao2025pr}, out-of-distribution generalization~\citep{jian2024tri,liu2025nested}, and anomaly detection~\citep{liu2025asynchronous,huang2024triadic}. \textit{Methodologically, our work utilizes bilevel optimization to explicitly decouple the conflicts between preference discrimination and foundational generative capabilities, thereby enhancing the stability and effectiveness of alignment training.}

\section{Conclusion}

We presented a general theoretical framework for understanding BT-style direct preference learning methods, including DPO and its several recent variants. By analyzing probability evolution under a unified objectives, we theoretically identified fundamental limitations shared across these methods. Motivated by these insights, we introduced a principled bilevel optimization framework, SPO, which integrates supervised fine-tuning and preference optimization within a single bilevel objective. Our empirical results validate the theoretical findings and show that SPO improves the stability of alignment training and enhances the performance of existing BT-style preference learning methods. Our work offers new insights into the mechanisms of direct preference learning, and we hope these findings inspire the development of more reliable, interpretable aligment methods in the future.

\clearpage
\section{Limitations}\label{Limitations and Future Work}
While the proposed SPO framework shows promising improvements in direct preference alignment for LLMs, it is important to acknowledge its limitations. Since resources are limited, the experiments presented in this paper primarily focus on smaller models. Although the results indicate that SPO outperforms DPO in terms of alignment and performance, further validation on larger models (like 72B parameter models) is necessary to fully understand its scalability and generalizability.

Moreover, the effectiveness of SPO relies on the quality of the SFT data used in the lower-level optimization. Since the feasible region $\mathfrak{R}$ is defined by the lower-level constraint, SPO assumes that the SFT data provides a robust manifold of foundational capabilities. Consequently, if the SFT data is of low quality, it may limit the benefits of SPO. Investigating how to dynamically refine the feasible region when SFT data is suboptimal remains an important direction for future research.

Finally, our theoretical analysis, as presented in Theorem~\ref{thm:prob_change}, focuses on the effects of single-step gradient updates. While this provides valuable insights into the immediate effects of BT-style preference optimization behaviors, it does not account for the complex evolution of probability distributions over multiple training steps. In multi-step scenarios, the tangent kernel term in Theorem~\ref{thm:prob_change} may change~\citep{jacot2018neural} over training steps, introducing additional complexity that our analysis does not fully capture. Exploring these long-term dynamics presents an exciting opportunity for future work, and we hope our findings encourage the research community to investigate multi-step behaviors to further enhance the stability and interpretability of direct preference learning methods.

\bibliography{main.bib}

\clearpage
\section{Proofs} \label{app:A}

\subsection{Proof of Theorem \ref{thm:prob_change}}\label{app:A.1}

For a single preference data point \((\boldsymbol{x}, \boldsymbol{y}_w, \boldsymbol{y}_l)\), recall that the unified loss function is defined as
\begin{equation}
\begin{aligned}
&\mathcal{L}_{\mathrm{unified}}(\boldsymbol{\theta}; \mathcal{D})
=-\log \sigma\!\Bigl(
\mathcal{R}_\theta(\boldsymbol{y}_w\mid\boldsymbol{x}) \\
&-\mathcal{R}_\theta(\boldsymbol{y}_l\mid\boldsymbol{x})
-\gamma\Bigr) + \Omega(\boldsymbol{x},\boldsymbol{y}_w,\boldsymbol{y}_l),
\end{aligned}
\end{equation}
During gradient descent, the model parameters are updated as
\begin{equation}
\boldsymbol{\theta} \leftarrow \boldsymbol{\theta} - \eta \nabla_{\boldsymbol{\theta}} \mathcal{L}_{\mathrm{unified}},
\end{equation}
where \(\eta\) is the learning rate, and \(\nabla_{\boldsymbol{\theta}} \mathcal{L}_{\mathrm{unified}}\) is the gradient of the unified loss function with respect to \(\boldsymbol{\theta}\). Our goal is to compute the probability changes
\begin{equation}
\begin{aligned}
\Delta \pi_{\boldsymbol{\theta}}(\boldsymbol{y}_w|\boldsymbol{x}) &= \pi_{\boldsymbol{\theta} + \Delta \boldsymbol{\theta}}(\boldsymbol{y}_w|\boldsymbol{x}) - \pi_{\boldsymbol{\theta}}(\boldsymbol{y}_w|\boldsymbol{x}), \\
\Delta \pi_{\boldsymbol{\theta}}(\boldsymbol{y}_l|\boldsymbol{x}) &= \pi_{\boldsymbol{\theta} + \Delta \boldsymbol{\theta}}(\boldsymbol{y}_l|\boldsymbol{x}) - \pi_{\boldsymbol{\theta}}(\boldsymbol{y}_l|\boldsymbol{x}).
\end{aligned}
\end{equation}
By using a Taylor expansion at the current parameter point \(\boldsymbol{\theta}\), we have
\begin{equation}
\begin{aligned}
& \pi_{\boldsymbol{\theta} + \Delta \boldsymbol{\theta}}(\boldsymbol{y}|\boldsymbol{x}) \\
&= \pi_{\boldsymbol{\theta}}(\boldsymbol{y}|\boldsymbol{x}) + \nabla_{\boldsymbol{\theta}} \pi_{\boldsymbol{\theta}}(\boldsymbol{y}|\boldsymbol{x}) \cdot \Delta \boldsymbol{\theta} + \mathcal{O}(\|\Delta \boldsymbol{\theta}\|^2) \\
&= \pi_{\boldsymbol{\theta}}(\boldsymbol{y}|\boldsymbol{x}) + \nabla_{\boldsymbol{\theta}} \pi_{\boldsymbol{\theta}}(\boldsymbol{y}|\boldsymbol{x}) \cdot \Delta \boldsymbol{\theta} + \mathcal{O}(\eta ^2\|\nabla_{\boldsymbol{\theta}} \mathcal{L}\|^2). 
\end{aligned}
\end{equation}
To ensure the theoretical validity of the remainder term \(\mathcal{O}(\|\Delta \boldsymbol{\theta}\|^2)\), we assume that the policy \(\pi_{\boldsymbol{\theta}}(\boldsymbol{y}|\boldsymbol{x})\) is \(L\)-smooth with respect to \(\boldsymbol{\theta}\). This smoothness property implies that the Hessian is bounded, thereby justifying the first-order approximation when the update step is sufficiently small. Given that the learning rate \(\eta\) is typically small (e.g., \(10^{-5}\)) and the gradient is bounded \citep{jentzen2020overallerroranalysistraining}, the second-order terms \(\mathcal{O}(\eta ^2\|\nabla_{\boldsymbol{\theta}} \mathcal{L}\|^2)\) become negligible compared to the first-order term. Therefore, the probability change can be accurately approximated as
\begin{equation}
\pi_{\boldsymbol{\theta} + \Delta \boldsymbol{\theta}}(\boldsymbol{y}|\boldsymbol{x}) \approx \pi_{\boldsymbol{\theta}}(\boldsymbol{y}|\boldsymbol{x}) + \nabla_{\boldsymbol{\theta}} \pi_{\boldsymbol{\theta}}(\boldsymbol{y}|\boldsymbol{x})^\top \Delta \boldsymbol{\theta}.
\end{equation}
% Since the learning rate \(\eta\) is small, and gradient is usually bounded \citep{jentzen2020overallerroranalysistraining} and we also use gradient clipping \citep{zhao2022penalizing} in practical algorithms. Thus, we can ignore the second-order terms \(\mathcal{O}(\eta^2 \|\nabla_{\boldsymbol{\theta}} \mathcal{L}\|^2)\), which are typically on the order of \(10^{-10}\) (e.g., \(\eta = 10^{-5}\), \(\|\nabla_{\boldsymbol{\theta}} \mathcal{L}\| \approx 1\)) compared to the first-order term's order of \(10^{-5}\). Therefore, we have
% \begin{equation}
% \pi_{\boldsymbol{\theta} + \Delta \boldsymbol{\theta}}(\boldsymbol{y}|\boldsymbol{x}) \approx \pi_{\boldsymbol{\theta}}(\boldsymbol{y}|\boldsymbol{x}) + \nabla_{\boldsymbol{\theta}} \pi_{\boldsymbol{\theta}}(\boldsymbol{y}|\boldsymbol{x}) \cdot \Delta \boldsymbol{\theta}.
% \end{equation}
Thus, we can obtain
\begin{equation} \label{eq:delta_pi}
\begin{aligned}
\Delta \pi_{\boldsymbol{\theta}}(\boldsymbol{y}|\boldsymbol{x}) 
&= \pi_{\boldsymbol{\theta} + \Delta \boldsymbol{\theta}}(\boldsymbol{y}|\boldsymbol{x}) - \pi_{\boldsymbol{\theta}}(\boldsymbol{y}|\boldsymbol{x}) \\
&= \nabla_{\boldsymbol{\theta}} \pi_{\boldsymbol{\theta}}(\boldsymbol{y}|\boldsymbol{x}) \cdot \Delta \boldsymbol{\theta}\\
&= \pi_{\boldsymbol{\theta}}(\boldsymbol{y}|\boldsymbol{x}) \nabla_{\boldsymbol{\theta}} \log \pi_{\boldsymbol{\theta}}(\boldsymbol{y}|\boldsymbol{x}) \cdot \Delta \boldsymbol{\theta}.
\end{aligned}
\end{equation}
Next, we compute the gradient of the loss. For a single sample, the unified loss function is \(\mathcal{L}_{\mathrm{unified}} = - \log \sigma(z)\), and its gradient is
\begin{equation}
\nabla_{\boldsymbol{\theta}} \mathcal{L}_{\mathrm{unified}} = - \frac{1}{\sigma(z)} \cdot \sigma'(z) \cdot \nabla_{\boldsymbol{\theta}} z,
\end{equation}
where \(z = \mathcal{R}_\theta(\boldsymbol{y}_w | \boldsymbol{x}) -\mathcal{R}_\theta(\boldsymbol{y}_l | \boldsymbol{x})-\gamma\), and \(\mathcal{R}_\theta(\boldsymbol{y}| \boldsymbol{x}) = a \log \pi_\theta(\boldsymbol{y} | \boldsymbol{x}) + b \log \pi_{\mathrm{ref}}(\boldsymbol{y}| \boldsymbol{x})\) is a general score function parameterized by coefficients \(a\) and \(b\), \(\gamma\) is a margin term. The derivative of the sigmoid function is
\begin{equation}
\sigma'(z) = \sigma(z) (1 - \sigma(z)).
\end{equation}
Thus, we have
\begin{equation} \label{eq:loss_gradient}
\begin{aligned}
\nabla_{\boldsymbol{\theta}} \mathcal{L}_{\mathrm{unified}} &= - \frac{\sigma(z) (1 - \sigma(z))}{\sigma(z)} \cdot \nabla_{\boldsymbol{\theta}} z \\
&= - (1 - \sigma(z)) \cdot \nabla_{\boldsymbol{\theta}} z.
\end{aligned}
\end{equation}
Since \(\pi_{\mathrm{ref}}\) is a fixed reference model, \(z\) does not depend on \(\boldsymbol{\theta}\), and the margin term \(\gamma\) is a constant that does not depend on the model parameters \(\theta\). So we have
\begin{equation}
\begin{aligned}
\nabla_{\boldsymbol{\theta}} z = \left( a_w \nabla_{\boldsymbol{\theta}} \log \pi_{\boldsymbol{\theta}}(\boldsymbol{y}_w|\boldsymbol{x}) - a_l \nabla_{\boldsymbol{\theta}} \log \pi_{\boldsymbol{\theta}}(\boldsymbol{y}_l|\boldsymbol{x}) \right).
\end{aligned}
\end{equation}
Substituting this into the loss gradient in Eq. (\ref{eq:loss_gradient}), we obtain
\begin{equation}
    \label{eq:dpo_gradient}
\begin{aligned}
&\nabla_{\boldsymbol{\theta}} \mathcal{L}_{\mathrm{unified}}= - (1 - \sigma(z)) \cdot \\
& \quad  \left( a_w \nabla_{\boldsymbol{\theta}} \log \pi_{\boldsymbol{\theta}}(\boldsymbol{y}_w|\boldsymbol{x}) - a_l \nabla_{\boldsymbol{\theta}} \log \pi_{\boldsymbol{\theta}}(\boldsymbol{y}_l|\boldsymbol{x}) \right).
\end{aligned}
\end{equation}
So the parameter update is
\begin{equation} \label{eq:delta_theta}
\begin{aligned}
\Delta \boldsymbol{\theta} &= - \eta (1 - \sigma(z)) \nabla_{\boldsymbol{\theta}} \mathcal{L}_{\mathrm{unified}} %\\
%&= - \eta (1 - \sigma(z)) \left( a_w \nabla_{\boldsymbol{\theta}} \log \pi_{\boldsymbol{\theta}}(\boldsymbol{y}_w|\boldsymbol{x}) - a_l \nabla_{\boldsymbol{\theta}} \log \pi_{\boldsymbol{\theta}}(\boldsymbol{y}_l|\boldsymbol{x}) \right).
\end{aligned}
\end{equation}
According to Eq. (\ref{eq:delta_pi}), for the preferred sample \(\boldsymbol{y}_w\), the probability change is
\begin{equation}\label{eq:delta_pi_w}
\Delta \pi_{\boldsymbol{\theta}}(\boldsymbol{y}_w|\boldsymbol{x}) \approx \pi_{\boldsymbol{\theta}}(\boldsymbol{y}_w|\boldsymbol{x}) \nabla_{\boldsymbol{\theta}} \log \pi_{\boldsymbol{\theta}}(\boldsymbol{y}_w|\boldsymbol{x}) \cdot \Delta \boldsymbol{\theta}.
\end{equation}
Combining with Eq. (\ref{eq:delta_theta}) and (\ref{eq:delta_pi_w}), we have
\begin{equation}
\begin{aligned}
&\Delta \pi_{\boldsymbol{\theta}}(\boldsymbol{y}_w|\boldsymbol{x}) \\
=& \pi_{\boldsymbol{\theta}}(\boldsymbol{y}_w|\boldsymbol{x}) \nabla_{\boldsymbol{\theta}} \log \pi_{\boldsymbol{\theta}}(\boldsymbol{y}_w|\boldsymbol{x}) \eta (1 - \sigma(z)) \cdot \\
& \quad \left( a_w \nabla_{\boldsymbol{\theta}} \log \pi_{\boldsymbol{\theta}}(\boldsymbol{y}_w|\boldsymbol{x}) - a_l \nabla_{\boldsymbol{\theta}} \log \pi_{\boldsymbol{\theta}}(\boldsymbol{y}_l|\boldsymbol{x}) \right)\\
=& \eta (1 - \sigma(z)) \pi_{\boldsymbol{\theta}}(\boldsymbol{y}_w|\boldsymbol{x}) \left( a_w^2 \|\nabla_{\boldsymbol{\theta}} \log \pi_{\boldsymbol{\theta}}(\boldsymbol{y}_w|\boldsymbol{x})\|^2 \right. \\
& \quad \left. - a_w a_l \nabla_{\boldsymbol{\theta}} \log \pi_{\boldsymbol{\theta}}(\boldsymbol{y}_w|\boldsymbol{x}) \cdot \nabla_{\boldsymbol{\theta}} \log \pi_{\boldsymbol{\theta}}(\boldsymbol{y}_l|\boldsymbol{x}) \right)
.
\end{aligned}
\end{equation}
Following a similar logic, we can obtain the probability change for the dispreferred sample \(y_l\). By settings \( \boldsymbol{g}_w = a_w \nabla_{\boldsymbol{\theta}} \log \pi_{\boldsymbol{\theta}}(\boldsymbol{y}_w|\boldsymbol{x})\) and \(\boldsymbol{g}_l = a_l \nabla_{\boldsymbol{\theta}} \log \pi_{\boldsymbol{\theta}}(\boldsymbol{y}_l|\boldsymbol{x})\), we complete the proof of Theorem \ref{thm:prob_change}.

\subsection{Proof of Corollary \ref{corollary:bound_on_probability_change}}\label{app:A.2}

According to Theorem \ref{thm:prob_change}, the probability change for the preferred sample \(\boldsymbol{y}_w\) is given by
\begin{equation}
\begin{aligned}
    % \resizebox{0.94\hsize}{!}{$
\Delta_w = &\eta (1-\sigma(z)) \pi_{\boldsymbol{\theta}} 
(\boldsymbol{y}_w|\boldsymbol{x})  a_w \nabla_{\boldsymbol{\theta}}\log \pi_{\boldsymbol{\theta}}(\boldsymbol{y}_w|\boldsymbol{x}) \cdot\\
&\left( a_w \nabla_{\boldsymbol{\theta}}\log \pi_{\boldsymbol{\theta}}(\boldsymbol{y}_w|\boldsymbol{x}) - a_l \nabla_{\boldsymbol{\theta}} \log \pi_{\boldsymbol{\theta}}(\boldsymbol{y}_l|\boldsymbol{x})\right).
% $}
\end{aligned}
\end{equation}

For simplification, we denote \(a_w \nabla_{\boldsymbol{\theta}} \log \pi_{\boldsymbol{\theta}}(\boldsymbol{y}_w|\boldsymbol{x})\) as \(\boldsymbol{g}_w\) and \(a_l \nabla_{\boldsymbol{\theta}} \log \pi_{\boldsymbol{\theta}}(\boldsymbol{y}_l|\boldsymbol{x})\) as \(\boldsymbol{g}_l\). By the Cauchy-Schwarz inequality, we have
\begin{equation}
\begin{aligned}
\boldsymbol{g}_w \cdot \boldsymbol{g}_l &\leq \|\boldsymbol{g}_w\| \|\boldsymbol{g}_l\|.
\end{aligned}
\end{equation}

Thus, we get
\begin{equation}
\begin{aligned}
\|\boldsymbol{g}_w\|^2 - \boldsymbol{g}_w \cdot \boldsymbol{g}_l &\geq \|\boldsymbol{g}_w\|^2 - \|\boldsymbol{g}_w\| \|\boldsymbol{g}_l\| \\
&= \|\boldsymbol{g}_w\| \left( \|\boldsymbol{g}_w\| - \|\boldsymbol{g}_l\| \right).
\end{aligned}
\end{equation}

Substituting this into the expression for \(\Delta \pi_{\boldsymbol{\theta}}(\boldsymbol{y}_w|\boldsymbol{x})\), we obtain
\begin{equation}
\begin{aligned}
\Delta_w &\geq \eta (1 - \sigma(z)) \pi_{\boldsymbol{\theta}}(\boldsymbol{y}_w|\boldsymbol{x}) \cdot \\
& \quad \|\boldsymbol{g}_w\| \left( \|\boldsymbol{g}_w\| - \|\boldsymbol{g}_l\| \right).
\end{aligned}
\end{equation}
This gives the lower bound for the probability change of the preferred sample \(\boldsymbol{y}_w\).

Similarly, for the dispreferred sample \(\boldsymbol{y}_l\), Theorem \ref{thm:prob_change} gives
\begin{equation}
\begin{aligned}
    % \resizebox{0.94\hsize}{!}{$
\Delta_l = &\eta (1-\sigma(z)) \pi_{\boldsymbol{\theta}} 
(\boldsymbol{y}_l|\boldsymbol{x}) \nabla_{\boldsymbol{\theta}}\log \pi_{\boldsymbol{\theta}}(\boldsymbol{y}_l|\boldsymbol{x}) \cdot\\
&\left(a_w \nabla_{\boldsymbol{\theta}}\log \pi_{\boldsymbol{\theta}}(\boldsymbol{y}_w|\boldsymbol{x}) - a_l \nabla_{\boldsymbol{\theta}} \log \pi_{\boldsymbol{\theta}}(\boldsymbol{y}_l|\boldsymbol{x})\right), 
% $}
\end{aligned}
\end{equation}
Applying the Cauchy-Schwarz inequality again, we have
\begin{equation}
\begin{aligned}
\boldsymbol{g}_l \cdot \boldsymbol{g}_w - \|\boldsymbol{g}_l\|^2 &\leq \|\boldsymbol{g}_l\| \|\boldsymbol{g}_w\| - \|\boldsymbol{g}_l\|^2 \\
&= \|\boldsymbol{g}_l\| \left( \|\boldsymbol{g}_w\| - \|\boldsymbol{g}_l\| \right).
\end{aligned}
\end{equation}

Substituting this into the expression for \(\Delta \pi_{\boldsymbol{\theta}}(\boldsymbol{y}_l|\boldsymbol{x})\), we obtain
\begin{equation}
\begin{aligned}
\Delta_l &\leq \eta (1 - \sigma(z)) \pi_{\boldsymbol{\theta}}(\boldsymbol{y}_l|\boldsymbol{x}) \cdot \\
& \quad \|\boldsymbol{g}_l\| \left( \|\boldsymbol{g}_w\| - \|\boldsymbol{g}_l\| \right).
\end{aligned}
\end{equation}

This completes the proof of Corollary \ref{corollary:bound_on_probability_change}.

\subsection{Proof of Corollary \ref{corollary:relative_change}}\label{app:A.3}

According to Theorem \ref{thm:prob_change}, the probability changes for the preferred and dispreferred samples are given by
\begin{equation}
\begin{aligned} 
\Delta_w &= \eta (1-\sigma(z)) \pi_{\boldsymbol{\theta}}(\boldsymbol{y}_w|\boldsymbol{x}) \boldsymbol{g}_w \cdot (\boldsymbol{g}_w - \boldsymbol{g}_l), \\
\Delta_l &= \eta (1-\sigma(z)) \pi_{\boldsymbol{\theta}}(\boldsymbol{y}_l|\boldsymbol{x}) \boldsymbol{g}_l \cdot (\boldsymbol{g}_w - \boldsymbol{g}_l).
\end{aligned}
\end{equation}
By denoting \(\Delta = \Delta_w - \Delta_l\), we have
\begin{equation}
\begin{aligned}
\Delta &= \eta (1 - \sigma(z)) \Big[ \pi_{\boldsymbol{\theta}}(\boldsymbol{y}_w|\boldsymbol{x}) \boldsymbol{g}_w \cdot (\boldsymbol{g}_w - \boldsymbol{g}_l) \\
& \quad - \pi_{\boldsymbol{\theta}}(\boldsymbol{y}_l|\boldsymbol{x}) \boldsymbol{g}_l \cdot (\boldsymbol{g}_w - \boldsymbol{g}_l) \Big] \\    
&= \eta (1 - \sigma(z)) \Big[ \pi_{\boldsymbol{\theta}}(\boldsymbol{y}_w|\boldsymbol{x}) \|\boldsymbol{g}_w\|^2 \\
&\quad + \pi_{\boldsymbol{\theta}}(\boldsymbol{y}_l|\boldsymbol{x}) \|\boldsymbol{g}_l\|^2 \\
& \quad - (\pi_{\boldsymbol{\theta}}(\boldsymbol{y}_w|\boldsymbol{x}) + \pi_{\boldsymbol{\theta}}(\boldsymbol{y}_l|\boldsymbol{x})) \boldsymbol{g}_w \cdot \boldsymbol{g}_l \Big].
\end{aligned}
\end{equation}
To prove the non-negativity of \(\Delta\), we analyze the term inside the brackets based on the sign of the inner product \(\boldsymbol{g}_w \cdot \boldsymbol{g}_l\):

\textbf{Case 1: \(\boldsymbol{g}_w \cdot \boldsymbol{g}_l \leq 0\).}
In this case, the preferred and dispreferred gradients are orthogonal or point in opposite directions. Since \(\pi_{\boldsymbol{\theta}}(\boldsymbol{y}_w|\boldsymbol{x}), \pi_{\boldsymbol{\theta}}(\boldsymbol{y}_l|\boldsymbol{x}) > 0\) and \(\eta(1-\sigma(z)) > 0\), the term \(-(\pi_{\boldsymbol{\theta}}(\boldsymbol{y}_w|\boldsymbol{x}) + \pi_{\boldsymbol{\theta}}(\boldsymbol{y}_l|\boldsymbol{x})) \boldsymbol{g}_w \cdot \boldsymbol{g}_l\) is non-negative. Therefore:
\begin{equation}
\begin{aligned}
\Delta &\geq \eta (1 - \sigma(z)) \bigl( \pi_{\boldsymbol{\theta}}(\boldsymbol{y}_w|\boldsymbol{x}) \|\boldsymbol{g}_w\|^2 \\
& \quad + \pi_{\boldsymbol{\theta}}(\boldsymbol{y}_l|\boldsymbol{x}) \|\boldsymbol{g}_l\|^2 \bigr) \geq 0.
\end{aligned}
\end{equation}

\textbf{Case 2: \(\boldsymbol{g}_w \cdot \boldsymbol{g}_l > 0\).}
This case represents the gradient geometric alignment scenario. By the AM-GM inequality, we have \(\pi_{\boldsymbol{\theta}}(\boldsymbol{y}_w|\boldsymbol{x}) + \pi_{\boldsymbol{\theta}}(\boldsymbol{y}_l|\boldsymbol{x}) \geq 2 \sqrt{\pi_{\boldsymbol{\theta}}(\boldsymbol{y}_w|\boldsymbol{x}) \pi_{\boldsymbol{\theta}}(\boldsymbol{y}_l|\boldsymbol{x})}\). Since \(\boldsymbol{g}_w \cdot \boldsymbol{g}_l > 0\), it follows that
\begin{equation} \label{eq:am_gm_application}
\begin{aligned}
&(\pi_{\boldsymbol{\theta}}(\boldsymbol{y}_w|\boldsymbol{x}) + \pi_{\boldsymbol{\theta}}(\boldsymbol{y}_l|\boldsymbol{x})) \boldsymbol{g}_w \cdot \boldsymbol{g}_l \\
&\geq 2 \sqrt{\pi_{\boldsymbol{\theta}}(\boldsymbol{y}_w|\boldsymbol{x}) \pi_{\boldsymbol{\theta}}(\boldsymbol{y}_l|\boldsymbol{x})} \boldsymbol{g}_w \cdot \boldsymbol{g}_l.
\end{aligned}
\end{equation}
Substituting Eq.~\eqref{eq:am_gm_application} into the expression for \(\Delta\), we obtain
\begin{equation}
    \resizebox{0.99\hsize}{!}{$
\begin{aligned}
\Delta &\geq \eta (1 - \sigma(z)) \Bigl[ \pi_{\boldsymbol{\theta}}(\boldsymbol{y}_w|\boldsymbol{x}) \|\boldsymbol{g}_w\|^2 + \pi_{\boldsymbol{\theta}}(\boldsymbol{y}_l|\boldsymbol{x}) \|\boldsymbol{g}_l\|^2 \\
& \quad - 2 \sqrt{\pi_{\boldsymbol{\theta}}(\boldsymbol{y}_w|\boldsymbol{x}) \pi_{\boldsymbol{\theta}}(\boldsymbol{y}_l|\boldsymbol{x})} \boldsymbol{g}_w \cdot \boldsymbol{g}_l \Bigr] \\
&= \eta (1 - \sigma(z)) \Bigl\| \sqrt{\pi_{\boldsymbol{\theta}}(\boldsymbol{y}_w|\boldsymbol{x})} \boldsymbol{g}_w - \sqrt{\pi_{\boldsymbol{\theta}}(\boldsymbol{y}_l|\boldsymbol{x})} \boldsymbol{g}_l \Bigr\|^2.
\end{aligned}
$}
\end{equation}
Since the squared norm is always non-negative, we conclude \(\Delta \geq 0\) in all cases. This completes the proof of Corollary \ref{corollary:relative_change}.

\subsection{Proof of Proposition \ref{prop:surrogate_objective}}\label{app:surrogate}

\begin{proposition}
    [Surrogate Objective for Gradient Constraint]\label{prop:surrogate_objective}
For each update step, the increase in log-probability for the preferred sample \(\boldsymbol{y}_w\) is lower-bounded by a term proportional to the difference in gradient norms:
\begin{equation}
\begin{aligned}
\log \pi_{\theta + \Delta \boldsymbol{\theta}}(\boldsymbol{y}_w | \boldsymbol{x}) - \log \pi_{\theta}(\boldsymbol{y}_w | \boldsymbol{x})
\geq \\
\eta (1 - \sigma(z)) \|\boldsymbol{g}_w\| \left( \|\boldsymbol{g}_w\| - \|\boldsymbol{g}_l\| \right).
\end{aligned}
\end{equation}
\end{proposition}

\noindent\textit{Proof.} To ensure the preferred sample \(\boldsymbol{y}_w\) receives a larger probability increase than the dispreferred sample \(\boldsymbol{y}_l\), we aim to satisfy the condition \(\|\boldsymbol{g}_w\| \ge \|\boldsymbol{g}_l\|\). However, directly constraining gradients is computationally expensive due to the implicit Hessian terms. 

For simplification, let \(\boldsymbol{g}_w = a_w \nabla_{\boldsymbol{\theta}} \log \pi_{\boldsymbol{\theta}}(\boldsymbol{y}_w|\boldsymbol{x})\) and \(\boldsymbol{g}_l = a_l \nabla_{\boldsymbol{\theta}} \log \pi_{\boldsymbol{\theta}}(\boldsymbol{y}_l|\boldsymbol{x})\). Under the assumption that \(\log \pi_{\boldsymbol{\theta}}(\boldsymbol{y}|\boldsymbol{x})\) is \(L\)-smooth, the first-order Taylor expansion yields:
\begin{equation}
\begin{aligned}
\log \pi_{\theta + \Delta \boldsymbol{\theta}}(\boldsymbol{y}_w | \boldsymbol{x}) - \log \pi_{\theta}(\boldsymbol{y}_w | \boldsymbol{x}) \approx \Delta \boldsymbol{\theta}^\top \boldsymbol{g_w}.
\end{aligned}
\end{equation}
Substituting the parameter update \(\Delta \boldsymbol{\theta} = \eta (1 - \sigma(z)) (\boldsymbol{g}_w - \boldsymbol{g}_l)\) into the above:
\begin{equation}
\begin{aligned}
&\log \pi_{\theta + \Delta \boldsymbol{\theta}}(\boldsymbol{y}_w | \boldsymbol{x}) - \log \pi_{\theta}(\boldsymbol{y}_w | \boldsymbol{x}) \\
&\approx \eta (1 - \sigma(z)) (\boldsymbol{g}_w - \boldsymbol{g}_l)^\top \boldsymbol{g}_w \\
&= \eta (1 - \sigma(z)) \left( \|\boldsymbol{g}_w\|^2 - \boldsymbol{g}_w^\top \boldsymbol{g}_l \right).
\end{aligned}
\end{equation}
By the Cauchy-Schwarz inequality, \(\boldsymbol{g}_w^\top \boldsymbol{g}_l \leq \|\boldsymbol{g}_w\| \|\boldsymbol{g}_l\|\), which implies:
\begin{equation}
\begin{aligned}
&\log \pi_{\theta + \Delta \boldsymbol{\theta}}(\boldsymbol{y}_w | \boldsymbol{x}) - \log \pi_{\theta}(\boldsymbol{y}_w | \boldsymbol{x}) \\
&\geq \eta (1 - \sigma(z)) \left( \|\boldsymbol{g}_w\|^2 - \|\boldsymbol{g}_w\| \|\boldsymbol{g}_l\| \right) \\
&= \eta (1 - \sigma(z)) \|\boldsymbol{g}_w\| \left( \|\boldsymbol{g}_w\| - \|\boldsymbol{g}_l\| \right).
\end{aligned}
\end{equation}
% This establishes that the log-probability of the preferred sample is guaranteed to increase if \(\|\boldsymbol{g}_w\| > \|\boldsymbol{g}_l\|\). Therefore, maintaining \(\log \pi_{\boldsymbol{\theta}}(\boldsymbol{y}_w|\boldsymbol{x})\) above its previous value serves as a tractable surrogate for the gradient norm constraint. To enforce thisduring training, we introduce the following regularization term
This result shows that the log-probability increase of the preferred sample is controlled by the difference in gradient norms: when the gradient norm of $\boldsymbol{y}_w$ dominates that of $\boldsymbol{y}_l$, the lower bound becomes larger. Therefore, instead of directly penalizing the gradient norm difference, we adopt the log-probability change as a tractable surrogate objective, which leads to the regularization term
\begin{equation}
\begin{aligned}
\mathcal{L}_{\text{reg}} = \max\left( 0, \log \pi_{\boldsymbol{\theta}}^{\text{old}}(\boldsymbol{y}_w|\boldsymbol{x}) - \log \pi_{\boldsymbol{\theta}}(\boldsymbol{y}_w|\boldsymbol{x}) \right).
\end{aligned}
\end{equation}
This completes the proof. This term penalizes any decrease in the log-probability of the preferred sample after an update. In practice, to stabilize training, we can use the log-probability from a previous checkpoint, an exponential moving average of model parameters as \(\log \pi_{\boldsymbol{\theta}}^{\text{old}}(\boldsymbol{y}_w|\boldsymbol{x})\), or even the log-probability from the reference model. In our experiments, we found using the reference model's log-probability works well. 

\subsection{Proof of Theorem \ref{thm:mass_shift}}\label{app:A.4}

For simplification, we denote \(a_w \nabla_{\boldsymbol{\theta}} \log \pi_{\boldsymbol{\theta}}(\boldsymbol{y}_w|\boldsymbol{x})\) as \(\boldsymbol{g}_w\) and \(a_l \nabla_{\boldsymbol{\theta}} \log \pi_{\boldsymbol{\theta}}(\boldsymbol{y}_l|\boldsymbol{x})\) as \(\boldsymbol{g}_l\). From Theorem~\ref{thm:prob_change}, for any output \( \boldsymbol{y} \in \mathcal{Y} \), the probability change after one gradient update is given by
\begin{equation}
\begin{aligned}
\Delta \pi_{\boldsymbol{\theta}}(\boldsymbol{y}|\boldsymbol{x}) &= \eta (1 - \sigma(z)) \pi_{\boldsymbol{\theta}}(\boldsymbol{y}|\boldsymbol{x}) \cdot \\
&\quad  \boldsymbol{g}_y \cdot \left( \boldsymbol{g}_w - \boldsymbol{g}_l \right).
\end{aligned}
\end{equation}

Thus, \(\Delta \pi_{\boldsymbol{\theta}}(\boldsymbol{y}|\boldsymbol{x})\) is determined by two factors:
\begin{itemize}
    \item The current probability \(\pi_{\boldsymbol{\theta}}(\boldsymbol{y}|\boldsymbol{x})\): A response with larger initial probability naturally contributes more to the total mass change.
    \item The gradient inner product \(\boldsymbol{g}_y \cdot (\boldsymbol{g}_w - \boldsymbol{g}_l)\): A positive inner product indicates that the output's gradient is more aligned with the preferred sample's gradient, leading to an increase in probability. Conversely, a negative inner product indicates alignment with the dispreferred sample's gradient, resulting in a decrease in probability.
\end{itemize}
Therefore, we can define the following four disjoint sets in the output space \( \mathcal{Y} \) based on these two factors:

(1) $\mathcal{D}_{sw}=\left\{
\boldsymbol{y}\in\mathcal{Y}:
\boldsymbol{g}_{y}^{\top}\boldsymbol{g}_w \gg
\boldsymbol{g}_{y}^{\top}\boldsymbol{g}_l\right\}.$ 

(2) $\mathcal{D}_{sl}=\left\{
\boldsymbol{y}\in\mathcal{Y}:
\boldsymbol{g}_{y}^{\top}\boldsymbol{g}_l \gg
\boldsymbol{g}_{y}^{\top}\boldsymbol{g}_w\right\}.$

(3) $\mathcal{D}_{*} = \{\boldsymbol{y}\in\mathcal{Y}\setminus (\mathcal{D}_{sw}\cup\mathcal{D}_{sl}): \pi_{\boldsymbol{\theta}}(\boldsymbol{y}|\boldsymbol{x}) \text{ is the top-k highest in } \mathcal{Y}\}.$

(4) $\mathcal{D}_o = \mathcal{Y}\setminus{\mathcal{D}_{sw}\cup\mathcal{D}_{sl}\cup\mathcal{D}_*}.$

The datasets \(\mathcal{D}_{sw}\) and \(\mathcal{D}_{sl}\) capture outputs whose gradients are strongly aligned with the preferred and dispreferred samples, respectively, while \(\mathcal{D}_*\) captures outputs with dominant probabilities that may not strongly align with either gradient, and \(\mathcal{D}_o\) includes all other outputs. 

According to the normalization condition of probabilities, we have
\begin{equation}
\begin{aligned}
&\sum_{\boldsymbol{y} \in \mathcal{D}_{sw}} \pi_{\boldsymbol{\theta}}(\boldsymbol{y}|\boldsymbol{x}) + \sum_{\boldsymbol{y} \in \mathcal{D}_{sl}} \pi_{\boldsymbol{\theta}}(\boldsymbol{y}|\boldsymbol{x}) \\
&+ \sum_{\boldsymbol{y} \in \mathcal{D}_{*}} \pi_{\boldsymbol{\theta}}(\boldsymbol{y}|\boldsymbol{x}) + \sum_{\boldsymbol{y} \in \mathcal{D}_{o}} \pi_{\boldsymbol{\theta}}(\boldsymbol{y}|\boldsymbol{x}) = 1.
\end{aligned}
\end{equation}

Next, we analyze two cases: (1) preferred sample probability decreases; (2) dispreferred sample probability decreases.

\subsubsection{Case 1: Preferred Sample Probability Decreases}

Suppose \( \Delta \pi_{\boldsymbol{\theta}}(\boldsymbol{y}_w|\boldsymbol{x}) < 0 \). By Theorem \ref{thm:prob_change}, the probability change is
\begin{equation}
\begin{aligned}
\Delta \pi_{\boldsymbol{\theta}}(\boldsymbol{y}_w|\boldsymbol{x}) &= \eta (1 - \sigma(z)) \pi_{\boldsymbol{\theta}}(\boldsymbol{y}_w|\boldsymbol{x}) \cdot \\
&\quad  \left( \|\boldsymbol{g}_w\|^2 - \boldsymbol{g}_w \cdot \boldsymbol{g}_l \right),
\end{aligned}
\end{equation}
When \( \Delta \pi_{\boldsymbol{\theta}}(\boldsymbol{y}_w|\boldsymbol{x}) < 0 \), we have
\begin{equation}
\begin{aligned}
\|\boldsymbol{g}_w\|^2 < \boldsymbol{g}_w \cdot \boldsymbol{g}_l.
\end{aligned}
\end{equation}
According to Corollary \ref{corollary:bound_on_probability_change}, the dispreferred sample’s probability change satisfies
\begin{equation}
\begin{aligned}
\Delta \pi_{\boldsymbol{\theta}}(\boldsymbol{y}_l|\boldsymbol{x}) &\leq \eta (1 - \sigma(z)) \pi_{\boldsymbol{\theta}}(\boldsymbol{y}_l|\boldsymbol{x}) \cdot \\
&\quad \|\boldsymbol{g}_l\| \left( \|\boldsymbol{g}_w\| - \|\boldsymbol{g}_l\| \right).
\end{aligned}
\end{equation}
If \( \|\boldsymbol{g}_w\| < \|\boldsymbol{g}_l\| \), then \( \Delta \pi_{\boldsymbol{\theta}}(\boldsymbol{y}_l|\boldsymbol{x}) < 0 \). According to Corollary \ref{corollary:relative_change}, we know that
\begin{equation}
\Delta \pi_{\boldsymbol{\theta}}(\boldsymbol{y}_w|\boldsymbol{x}) - \Delta \pi_{\boldsymbol{\theta}}(\boldsymbol{y}_l|\boldsymbol{x}) \geq 0,
\end{equation}
implying \( \Delta \pi_{\boldsymbol{\theta}}(\boldsymbol{y}_l|\boldsymbol{x}) \leq \Delta \pi_{\boldsymbol{\theta}}(\boldsymbol{y}_w|\boldsymbol{x}) < 0 \), so \( \Delta \pi_{\boldsymbol{\theta}}(\boldsymbol{y}_l|\boldsymbol{x}) < 0 \). By probability conservation, 

\begin{equation} \label{eq:prob_conservation}
\begin{aligned}
&\Delta_{sw} + \Delta_{sl} + \Delta_{*} + \Delta_{o} = 0,
\end{aligned}
\end{equation}
where \( \Delta_{sw} = \sum_{\boldsymbol{y} \in \mathcal{D}_{sw}} \Delta \pi_{\boldsymbol{\theta}}(\boldsymbol{y}|\boldsymbol{x}) \), \( \Delta_{sl} = \sum_{\boldsymbol{y} \in \mathcal{D}_{sl}} \Delta \pi_{\boldsymbol{\theta}}(\boldsymbol{y}|\boldsymbol{x}) \), \( \Delta_{*} = \sum_{\boldsymbol{y} \in \mathcal{D}_{*}} \Delta \pi_{\boldsymbol{\theta}}(\boldsymbol{y}|\boldsymbol{x}) \), and \( \Delta_{o} = \sum_{\boldsymbol{y} \in \mathcal{D}_{o}} \Delta \pi_{\boldsymbol{\theta}}(\boldsymbol{y}|\boldsymbol{x}) \). Since \( \Delta \pi_{\boldsymbol{\theta}}(\boldsymbol{y}_w|\boldsymbol{x}) < 0 \) and \( \Delta \pi_{\boldsymbol{\theta}}(\boldsymbol{y}_l|\boldsymbol{x}) < 0 \), so we have \( \Delta_{sw} < 0 \) and \( \Delta_{sl} < 0 \). Therefore, to satisfy the probability conservation in Eq.~\eqref{eq:prob_conservation}, at least one of \( \Delta_{*} \) or \( \Delta_{o} \) must be positive. Thus, we have
\begin{equation}
\begin{aligned}
\Delta_{*} + \Delta_{o} > 0.
\end{aligned}
\end{equation}
For any output \( \boldsymbol{y} \in \mathcal{Y} \), Theorem \ref{thm:prob_change} gives
\begin{equation}
\begin{aligned}
\Delta \pi_{\boldsymbol{\theta}}(\boldsymbol{y}|\boldsymbol{x}) &= \eta a (1 - \sigma(z)) \pi_{\boldsymbol{\theta}}(\boldsymbol{y}|\boldsymbol{x}) \boldsymbol{g} \cdot (\boldsymbol{g}_w - \boldsymbol{g}_l) \\
\end{aligned}
\end{equation}
So for the remaining outputs, we have
\begin{equation}
\begin{aligned}
\Delta_{o}
&= \sum_{\boldsymbol{y} \in \mathcal{D}_{*}} \Delta \pi_{\boldsymbol{\theta}}(\boldsymbol{y}|\boldsymbol{x}) + \sum_{\boldsymbol{y} \in \mathcal{D}_{o}} \Delta \pi_{\boldsymbol{\theta}}(\boldsymbol{y}|\boldsymbol{x}) \\
&= \eta (1 - \sigma(z)) \cdot \\
&\quad \sum_{\boldsymbol{y} \in \mathcal{D}_{*} \cup \mathcal{D}_{o}} \pi_{\boldsymbol{\theta}}(\boldsymbol{y}|\boldsymbol{x}) \boldsymbol{g} \cdot (\boldsymbol{g}_w - \boldsymbol{g}_l).
\end{aligned}
\end{equation}
Since the remaining outputs do not strongly align with either \( \boldsymbol{g}_w \) or \( \boldsymbol{g}_l \) and \(\boldsymbol{y}\in\mathcal{D}_{o}\) is out of the top \(k\) predictions, we have \( \boldsymbol{g} \cdot (\boldsymbol{g}_w - \boldsymbol{g}_l) \approx 0 \) for \( \boldsymbol{y} \in \mathcal{D}_{o} \) and \( \boldsymbol{g} \cdot (\boldsymbol{g}_w - \boldsymbol{g}_l) > 0 \) for \( \boldsymbol{y} \in \mathcal{D}_{*} \). Thus, the changes for all other outputs are negligible, i.e., \(\Delta_{o} \approx 0\). Therefore, we have
\begin{equation}
\begin{aligned}
\Delta_{*} \approx -(\Delta_{sw} + \Delta_{sl}) > 0.
\end{aligned}
\end{equation}
Hence, probability mass primarily flows to the high confidence outputs.

\subsubsection{Case 2: Dispreferred Sample Probability Decreases}
Suppose \( \Delta \pi_{\boldsymbol{\theta}}(\boldsymbol{y}_l|\boldsymbol{x}) < 0 \). By Theorem \ref{thm:prob_change}
\begin{equation}
\begin{aligned}
\Delta \pi_{\boldsymbol{\theta}}(\boldsymbol{y}_l|\boldsymbol{x}) &= \eta (1 - \sigma(z)) \pi_{\boldsymbol{\theta}}(\boldsymbol{y}_l|\boldsymbol{x}) \cdot \\ 
&\quad \left( \boldsymbol{g}_l \cdot \boldsymbol{g}_w - \|\boldsymbol{g}_l\|^2 \right),
\end{aligned}
\end{equation}
When \( \Delta \pi_{\boldsymbol{\theta}}(\boldsymbol{y}_l|\boldsymbol{x}) < 0 \), we have
\begin{equation}
\begin{aligned}
\boldsymbol{g}_l \cdot \boldsymbol{g}_w < \|\boldsymbol{g}_l\|^2.
\end{aligned}
\end{equation}
From Eq.~\eqref{eq:prob_conservation}, we have
\begin{equation}
\begin{aligned}
&\Delta_{sw} + \Delta_{*} + \Delta_{o} = -\Delta_{sl} > 0.
\end{aligned}
\end{equation}
Since the \(\Delta_{o} \approx 0\), we get
\begin{equation}
\begin{aligned}
&\Delta_{sw} + \Delta_{*} \approx -\Delta_{sl} > 0.
\end{aligned}
\end{equation}
Recall that the probability changes for \( \boldsymbol{y}_w \) and \( \boldsymbol{y}^* \) are
\begin{equation}
\begin{aligned}
\Delta \pi_{\boldsymbol{\theta}}(\boldsymbol{y}_w|\boldsymbol{x}) &= \eta (1 - \sigma(z)) \pi_{\boldsymbol{\theta}}(\boldsymbol{y}_w|\boldsymbol{x}) \cdot \\
&\quad \left( \|\boldsymbol{g}_w\|^2 - \boldsymbol{g}_w \cdot \boldsymbol{g}_l \right), \\
\Delta \pi_{\boldsymbol{\theta}}(\boldsymbol{y}^*|\boldsymbol{x}) &= \eta (1 - \sigma(z)) \pi_{\boldsymbol{\theta}}(\boldsymbol{y}^*|\boldsymbol{x}) \cdot \\
&\quad \left( \|\boldsymbol{g}^*\|^2 - \boldsymbol{g}^* \cdot \boldsymbol{g}_l \right).
\end{aligned}
\end{equation}
The flow of probability mass depends on the initial probabilities \( \pi_{\boldsymbol{\theta}}(\boldsymbol{y}_w|\boldsymbol{x}) \), \( \pi_{\boldsymbol{\theta}}(\boldsymbol{y}^*|\boldsymbol{x}) \), and gradient similarities. If \( \pi_{\boldsymbol{\theta}}(\boldsymbol{y}_w|\boldsymbol{x}) > \pi_{\boldsymbol{\theta}}(\boldsymbol{y}^*|\boldsymbol{x}) \) and \(\|\boldsymbol{g}_w\|^2 > \boldsymbol{g}_w \cdot \boldsymbol{g}_l\), then \( \Delta \pi_{\boldsymbol{\theta}}(\boldsymbol{y}_w|\boldsymbol{x}) > 0 \), and probability mass primarily flows to \( \boldsymbol{y}_w \). Otherwise, if \( \pi_{\boldsymbol{\theta}}(\boldsymbol{y}^*|\boldsymbol{x}) > \pi_{\boldsymbol{\theta}}(\boldsymbol{y}_w|\boldsymbol{x}) \) or the gradient similarity favors \( y^* \), then \( \Delta \pi_{\boldsymbol{\theta}}(\boldsymbol{y}^*|\boldsymbol{x}) > 0\). This completes the proof of Theorem \ref{thm:mass_shift}.

\subsection{Derivation of Closed-Form Gradient}\label{app:A.5}

We derive the gradient of the optimization objective in Eq.~\eqref{eq:penalized_opt}, defined as
% \begin{equation} \label{eq:objective}
% \min_{\boldsymbol{\theta}} \quad \mathcal{L}_{\mathrm{spo}}(\boldsymbol{x}, \boldsymbol{y}_w, \boldsymbol{y}_l; \boldsymbol{\theta}) := \mathcal{L}_{\mathrm{pref}}(\boldsymbol{x}, \boldsymbol{y}_w, \boldsymbol{y}_l; \boldsymbol{\theta}) + \lambda \left( \mathcal{L}_{\mathrm{sft}}(\boldsymbol{x}, \boldsymbol{y}_w; \boldsymbol{\theta}) - \mathcal{L}_{\mathrm{sft}}^{*}(\boldsymbol{x}, \boldsymbol{y}_w) \right), 
% \end{equation}
\begin{equation} \label{eq:objective}
\begin{aligned}
\min_{\boldsymbol{\theta}}\  &\mathcal{L}_{\mathrm{spo}}(\boldsymbol{x}, \boldsymbol{y}_w, \boldsymbol{y}_l; \boldsymbol{\theta}) := \mathcal{L}_{\mathrm{pref}}(\boldsymbol{x}, \boldsymbol{y}_w, \boldsymbol{y}_l; \boldsymbol{\theta}) \\
& + \lambda \left( \mathcal{L}_{\mathrm{sft}}(\boldsymbol{x}, \boldsymbol{y}_w; \boldsymbol{\theta}) - \mathcal{L}_{\mathrm{sft}}^{*}(\boldsymbol{x}, \boldsymbol{y}_w) \right)\\
& + \mathcal{L}_{\text{reg}}(\boldsymbol{x}, \boldsymbol{y}_w, \boldsymbol{y}_l; \boldsymbol{\theta}).
\end{aligned}
\end{equation}
where \(\mathcal{L}_{\mathrm{pref}}(\boldsymbol{x}, \boldsymbol{y}_w, \boldsymbol{y}_l; \boldsymbol{\theta}) = \mathcal{L}(\boldsymbol{x}, \boldsymbol{y}_w, \boldsymbol{y}_l; \boldsymbol{\theta}) + \gamma \mathcal{L}_{\text{reg}}(\boldsymbol{x}, \boldsymbol{y}_w; \boldsymbol{\theta})\).

For a single sample \((\boldsymbol{x}, \boldsymbol{y}_w, \boldsymbol{y}_l) \in \mathcal{D}_{\mathrm{tr}}\), where \(\boldsymbol{y}_w\) is the preferred response and \(\boldsymbol{y}_l\) is the dispreferred response, and \((\boldsymbol{x}, \boldsymbol{y}_w) \in \mathcal{D}_{\mathrm{sft}}\), we next derive the closed-form gradient \(\nabla_{\boldsymbol{\theta}} \mathcal{L}_{\mathrm{spo}}\).

For the objective $\mathcal{L}_{\mathrm{pref}}(\boldsymbol{\theta})$, the gradient is
\begin{equation}
\begin{aligned} \label{eq:grad_theta}
&\nabla_{\boldsymbol{\theta}} \mathcal{L}_{\mathrm{pref}} \\
&= \nabla_{\boldsymbol{\theta}} \mathcal{L}(\boldsymbol{x}, \boldsymbol{y}_w, \boldsymbol{y}_l; \boldsymbol{\theta}) + \gamma \nabla_{\boldsymbol{\theta}} \mathcal{L}_{\text{reg}}(\boldsymbol{x}, \boldsymbol{y}_w, \boldsymbol{y}_l; \boldsymbol{\theta}) \\
&\quad + \lambda \left( \nabla_{\boldsymbol{\theta}} \mathcal{L}_{\mathrm{sft}}(\boldsymbol{x}, \boldsymbol{y}_w; \boldsymbol{\theta}) - \nabla_{\boldsymbol{\theta}} \mathcal{L}_{\mathrm{sft}}^{*}(\boldsymbol{x}, \boldsymbol{y}_w; \boldsymbol{\theta}) \right).
\end{aligned}
\end{equation}
\paragraph{Preference Loss Gradient.} The preference learning objective for a single sample is
\begin{equation}
\mathcal{L}(\boldsymbol{x}, \boldsymbol{y}_w, \boldsymbol{y}_l; \boldsymbol{\theta}) = -\log \sigma(z),
\end{equation}
where \( z = \mathcal{R}_\theta(\boldsymbol{y}_w\mid\boldsymbol{x}) - \mathcal{R}_\theta(\boldsymbol{y}_l\mid\boldsymbol{x}) \). The gradient of \(\mathcal{L}(\boldsymbol{x}, \boldsymbol{y}_w, \boldsymbol{y}_l; \boldsymbol{\theta})\) is
\begin{equation}
\nabla_{\boldsymbol{\theta}} \mathcal{L}(\boldsymbol{x}, \boldsymbol{y}_w, \boldsymbol{y}_l; \boldsymbol{\theta}) = -\frac{1}{\sigma(z)} \sigma'(z) \nabla_{\boldsymbol{\theta}} z.
\end{equation}
Since \(\sigma'(z) = \sigma(z)(1 - \sigma(z))\), we have
\begin{equation}
\nabla_{\boldsymbol{\theta}} \mathcal{L}(\boldsymbol{x}, \boldsymbol{y}_w, \boldsymbol{y}_l; \boldsymbol{\theta}) = -(1 - \sigma(z)) \nabla_{\boldsymbol{\theta}} z.
\end{equation}
% Since \(
% \nabla_{\boldsymbol{\theta}} z = \beta \left( \nabla_{\boldsymbol{\theta}} \log \pi_{\boldsymbol{\theta}}(\boldsymbol{y}_w|\boldsymbol{x}) - \nabla_{\boldsymbol{\theta}} \log \pi_{\boldsymbol{\theta}}(\boldsymbol{y}_l|\boldsymbol{x}) \right)\) and \(\pi_{\mathrm{ref}}\) is independent of \(\boldsymbol{\theta}\). So we have
Since \(\pi_{\mathrm{ref}}\) is independent of \(\boldsymbol{\theta}\). So we have
\begin{equation} \label{eq:dpo_grad_theta}
\begin{aligned}
&\nabla_{\boldsymbol{\theta}} \mathcal{L}(\boldsymbol{x}, \boldsymbol{y}_w, \boldsymbol{y}_l; \boldsymbol{\theta}) = -a (1 - \sigma(z)) \cdot \\
&\quad \left( \nabla_{\boldsymbol{\theta}} \log \pi_{\boldsymbol{\theta}}(\boldsymbol{y}_w|\boldsymbol{x}) - \nabla_{\boldsymbol{\theta}} \log \pi_{\boldsymbol{\theta}}(\boldsymbol{y}_l|\boldsymbol{x}) \right).
\end{aligned}
\end{equation}

\paragraph{SFT Loss Gradient.} The gradient of SFT loss for a single sample is
\begin{equation} \label{eq:sft_grad_theta}
\nabla_{\boldsymbol{\theta}} \mathcal{L}_{\mathrm{sft}}(\boldsymbol{x}, \boldsymbol{y}_w; \boldsymbol{\theta}) = -\nabla_{\boldsymbol{\theta}} \log \pi_{\boldsymbol{\theta}}(\boldsymbol{y}_w|\boldsymbol{x}).
\end{equation}

\paragraph{Optimal SFT Loss Gradient.} The optimal SFT loss is
\begin{equation}
\mathcal{L}_{\mathrm{sft}}^{*}(\boldsymbol{x}, \boldsymbol{y}_w; \boldsymbol{\theta}) = \mathcal{L}_{\mathrm{sft}}(\boldsymbol{x}, \boldsymbol{y}_w; \boldsymbol{\theta}^{*}),
\end{equation}
where \(\boldsymbol{\theta}^{*} = \arg\min \mathcal{L}_{\mathrm{sft}}(\boldsymbol{x}, \boldsymbol{y}_w; \boldsymbol{\theta}^{\prime})\). This gradient is dependent on the optimal solution of the lower-level optimization problem.
According to ~\citep{nouiehed2019solving} Lemma A.5, the optimal SFT loss gradient can be evaluated directly at any optimal solution of the lower-level optimization problem under some mild assumptions. Based on this, we have
\begin{equation}
\begin{aligned}
\nabla_{\boldsymbol{\theta}} \mathcal{L}_{\mathrm{sft}}^{*}(\boldsymbol{x}, \boldsymbol{y}_w; \boldsymbol{\theta}) &= -\nabla_{\boldsymbol{\theta}} \log \pi_{\boldsymbol{\theta}^{*}}(\boldsymbol{y}_w|\boldsymbol{x}) \\
& \approx -\nabla_{\boldsymbol{\theta}} \log \pi_{\boldsymbol{\theta}^{(K-1)}}(\boldsymbol{y}_w|\boldsymbol{x}), 
\end{aligned}
\end{equation}
where $\boldsymbol{\theta}^{(K-1)}$ is the approximate solution of the lower-level optimization problem obtained after K-step gradient descent.

\paragraph{Regularization Loss Gradient.} Since we use the reference model as the initial model, we have \(\pi_{\boldsymbol{\theta}}^{\text{old}} = \pi_{\mathrm{ref}}\). So the regularization term can be rewritten as
\begin{equation}
\begin{aligned}
&\mathcal{L}_{\text{reg}}(\boldsymbol{x}, \boldsymbol{y}_w, \boldsymbol{y}_l; \boldsymbol{\theta}) \\
&= \max\big(0, \log \pi_{\text{ref}}(\boldsymbol{y}_w|\boldsymbol{x}) - \log \pi_{\boldsymbol{\theta}}(\boldsymbol{y}_w|\boldsymbol{x})\big).
\end{aligned}
\end{equation}
So the gradient of the regularization term is
\begin{equation} \label{eq:reg_grad_theta}
\begin{aligned}
&\nabla_{\boldsymbol{\theta}} \mathcal{L}_{\text{reg}}(\boldsymbol{x}, \boldsymbol{y}_w, \boldsymbol{y}_l; \boldsymbol{\theta}) = \\
&\begin{cases}
-\nabla_{\boldsymbol{\theta}} \log \pi_{\boldsymbol{\theta}}(\boldsymbol{y}_w|\boldsymbol{x}), & \text{if } t > 0, \\
0, & \text{otherwise}.
\end{cases}
\end{aligned}
\end{equation}
where \( t = \log \pi_{\text{ref}}(\boldsymbol{y}_w|\boldsymbol{x}) - \log \pi_{\boldsymbol{\theta}}(\boldsymbol{y}_w|\boldsymbol{x}) \). By substituting Eqs.~\eqref{eq:dpo_grad_theta}, \eqref{eq:sft_grad_theta}, \eqref{eq:reg_grad_theta} into Eq.~\eqref{eq:grad_theta}, we obtain the final gradient expression for \(\nabla_{\boldsymbol{\theta}} \mathcal{L}_{\mathrm{spo}}\). %The detailed formula is
% \begin{equation}
% \begin{aligned}
% \nabla_{\boldsymbol{\theta}} \mathcal{L}_{\mathrm{spo}} = & -(\beta (1 - \sigma(z)) + \gamma) \nabla_{\boldsymbol{\theta}} \log \pi_{\boldsymbol{\theta}}(\boldsymbol{y}_w|\boldsymbol{x}) + \beta (1 - \sigma(z)) \nabla_{\boldsymbol{\theta}} \log \pi_{\boldsymbol{\theta}}(\boldsymbol{y}_l|\boldsymbol{x}) \\
% & - \lambda \left( \nabla_{\boldsymbol{\theta}} \log \pi_{\boldsymbol{\theta}}(\boldsymbol{y}_w|\boldsymbol{x}) - \nabla_{\boldsymbol{\theta}} \log \pi_{\boldsymbol{\theta}^{(K-1)}}(\boldsymbol{y}_w|\boldsymbol{x}) \right), \\
% & \text{if } \log \pi_{\boldsymbol{\theta}}^{\text{ref}}(\boldsymbol{y}_w|\boldsymbol{x}) - \log \pi_{\boldsymbol{\theta}}(\boldsymbol{y}_w|\boldsymbol{x}) > 0; \\
% \nabla_{\boldsymbol{\theta}} \mathcal{L}_{\mathrm{spo}} = & -\beta (1 - \sigma(z)) \nabla_{\boldsymbol{\theta}} \log \pi_{\boldsymbol{\theta}}(\boldsymbol{y}_w|\boldsymbol{x}) + \beta (1 - \sigma(z)) \nabla_{\boldsymbol{\theta}} \log \pi_{\boldsymbol{\theta}}(\boldsymbol{y}_l|\boldsymbol{x}) \\
% & - \lambda \left( \nabla_{\boldsymbol{\theta}} \log \pi_{\boldsymbol{\theta}}(\boldsymbol{y}_w|\boldsymbol{x}) - \nabla_{\boldsymbol{\theta}} \log \pi_{\boldsymbol{\theta}^{(K-1)}}(\boldsymbol{y}_w|\boldsymbol{x}) \right), \\
% & \text{otherwise}.
% \end{aligned}
% \end{equation}  

\subsection{Quantitative Analysis of Probability Change for $y^*$} \label{app:probability_change_y_star}

To address the mass allocation issue raised in Theorem \ref{thm:mass_shift}, we derive the quantitative change for an arbitrary output $\boldsymbol{y}^* \in \mathcal{D}_*$. Following the Taylor expansion framework in Appendix \ref{app:A.1}, the probability change $\Delta \pi_{\boldsymbol{\theta}}(\boldsymbol{y}^*|\boldsymbol{x})$ is
\begin{equation} \label{eq:delta_pi_star_split}
\begin{aligned}
\Delta \pi_{\boldsymbol{\theta}}(\boldsymbol{y}^*|\boldsymbol{x}) 
&\approx \pi_{\boldsymbol{\theta}}(\boldsymbol{y}^*|\boldsymbol{x}) \nabla_{\boldsymbol{\theta}} \log \pi_{\boldsymbol{\theta}}(\boldsymbol{y}^*|\boldsymbol{x})^\top \Delta \boldsymbol{\theta} \\
&= \eta (1 - \sigma(z)) \pi_{\boldsymbol{\theta}}(\boldsymbol{y}^*|\boldsymbol{x}) \boldsymbol{g}_{y^*}^\top \\
& \quad \cdot \left( a_w \boldsymbol{g}_w - a_l \boldsymbol{g}_l \right),
\end{aligned}
\end{equation}
where $\boldsymbol{g}_y = \nabla_{\boldsymbol{\theta}} \log \pi_{\boldsymbol{\theta}}(\boldsymbol{y}|\boldsymbol{x})$. Setting $a_w = a_l = \beta$ (as in standard DPO), we obtain the final quantitative form
\begin{equation} \label{eq:y_star_final_narrow}
\Delta \pi_{\boldsymbol{\theta}}(\boldsymbol{y}^*|\boldsymbol{x}) \approx \eta \beta (1 - \sigma(z)) \pi_{\boldsymbol{\theta}}(\boldsymbol{y}^*|\boldsymbol{x}) \boldsymbol{g}_{y^*}^\top \left( \boldsymbol{g}_w - \boldsymbol{g}_l \right).
\end{equation}

%%%%%%%%%%%%%%%%%%%%%%%%%%%%%%%%%%%%%%%%%%%%%%%%%%%%%%%%%%%%

\section{BT-style Preference Learning Methods}\label{app:BT Model-based Direct Preference Learning Methods}

\paragraph{CPO.} Contrastive Preference Optimization~\citep{xu2024contrastive} addresses the limitations of traditional supervised fine-tuning by optimizing a model to directly prefer higher-quality outputs over lower-quality ones, the CPO objective is defined as:
\begin{equation}
\begin{aligned}
&\mathcal{L}_{\mathrm{cpo}}(\boldsymbol{\theta}; \mathcal{D})
= -\log \sigma \Bigl(
\beta \log \pi_\theta(\boldsymbol{y}_w \mid \boldsymbol{x})
\\
&- \beta \log \pi_\theta(\boldsymbol{y}_l \mid \boldsymbol{x})
\Bigr) - \lambda \log \pi_\theta(\boldsymbol{y}_w \mid \boldsymbol{x}),
\end{aligned}
\end{equation}
where $\lambda$ is a regularization hyperparameter.

\paragraph{R-DPO.} R-DPO~\citep{park2024disentangling} extends the standard DPO framework by introducing an explicit length regularization term to mitigate the model's tendency to over-prefer verbose responses. The R-DPO objective is formulated as
\begin{equation}
\begin{aligned}
&\mathcal{L}_{\mathrm{rdpo}}(\boldsymbol{\theta}; \mathcal{D}) =
-\log \sigma\!\Bigl(
\beta \log \tfrac{\pi_\theta(\boldsymbol{y}_w \mid \boldsymbol{x})}{\pi_{\mathrm{ref}}(\boldsymbol{y}_w \mid \boldsymbol{x})}\\
&-\beta \log \tfrac{\pi_\theta(\boldsymbol{y}_l \mid \boldsymbol{x})}{\pi_{\mathrm{ref}}(\boldsymbol{y}_l \mid \boldsymbol{x})}
+ \alpha\left(|\boldsymbol{y}_w| - |\boldsymbol{y}_l|\right)
\Bigr),
\end{aligned}
\end{equation}
where $\alpha$ is the strength of the length regularization and $|\boldsymbol{y}|$ denotes the token length of a response.

\paragraph{DPOP.} DPOP~\citep{pal2024smaug} mitigates a failure mode of standard DPO in which the log-likelihood of the preferred completion may be inadvertently reduced by adding a penalty term that prevents the preferred response from falling below its reference-model likelihood. The objective is defined as
\begin{equation}
\begin{aligned}
&\mathcal{L}_{\mathrm{dpop}}(\boldsymbol{\theta})
= - \log \sigma\!\Bigl(
\beta\!\Bigl(
\log \tfrac{\pi_\theta(\boldsymbol{y}_w\mid\boldsymbol{x})}
         {\pi_{\mathrm{ref}}(\boldsymbol{y}_w\mid\boldsymbol{x})}\\
&-\log \tfrac{\pi_\theta(\boldsymbol{y}_l\mid\boldsymbol{x})}
         {\pi_{\mathrm{ref}}(\boldsymbol{y}_l\mid\boldsymbol{x})}
\Bigr)
-\,\lambda \cdot \max\!\left(0,\,
\log \tfrac{\pi_{\mathrm{ref}}(\boldsymbol{y}_w\mid\boldsymbol{x})}
         {\pi_\theta(\boldsymbol{y}_w\mid\boldsymbol{x})}
\right)
\Bigr),
\end{aligned}
\end{equation}
where $\lambda>0$ controls the strength of the positive constraint.

% \paragraph{SPL.} Soft Preference Learning~\citep{sharifnassab2024soft} generalizes standard DPO by decoupling the KL regularization term into a separate entropy bonus and cross-entropy penalty, which enables independent control over output diversity and the strength of the reference-model prior. The DPO-style objective is given by
% \begin{equation}
%     \begin{aligned}
% \label{eq:spl_dpo}
% \mathcal{L}_{\mathrm{spl}}(\theta)
% =
% -\log\sigma\!&\Bigl(
% \alpha\log\tfrac{\pi_\theta(\boldsymbol{y}_w\mid \boldsymbol{x})}{\pi_\theta(\boldsymbol{y}_l\mid \boldsymbol{x})}\\
% &-
% \beta\log\tfrac{\pi_{\mathrm{ref}}(\boldsymbol{y}_w\mid \boldsymbol{x})}{\pi_{\mathrm{ref}}(\boldsymbol{y}_l\mid \boldsymbol{x})}
% \Bigr).
%     \end{aligned}
% \end{equation}
% where $\alpha, \beta > 0$ controls the strength of the entropy bonus and cross-entropy penalty, respectively. 

\paragraph{SimPO.} Simple Preference Optimization~\citep{meng2024simpo} improves standard DPO by using a reference-free, length-normalized reward based on average log-likelihood, which better aligns with human preference and avoids length bias. The resulting objective is
\begin{equation}
\begin{aligned}
\mathcal{L}_{\mathrm{simpo}}(\boldsymbol{\theta}; \mathcal{D})
&=
-\log \sigma \!\Bigl(
\frac{\beta}{|\boldsymbol{y}_w|} \log \pi_\theta(\boldsymbol{y}_w \mid \boldsymbol{x})\\
&-
\frac{\beta}{|\boldsymbol{y}_l|} \log \pi_\theta(\boldsymbol{y}_l \mid \boldsymbol{x})
-
\gamma
\Bigr),
\end{aligned}
\end{equation}
where $\beta$ scales the reward difference, $|\boldsymbol{y}|$ denotes the response length, and $\gamma > 0$ is a target reward margin that ensures a minimum reward gap between preferred and dispreferred responses.

\paragraph{AlphaDPO.} AlphaDPO extends standard preference optimization by introducing an adaptive, instance-dependent reward margin that adjusts the separation between preferred and dispreferred responses. The resulting objective is
\begin{equation}
\begin{aligned}
&\mathcal{L}_{\mathrm{alphaDPO}}(\boldsymbol{\theta}; \mathcal{D})
=
-\log \sigma \!\Bigl(
\frac{\beta}{|\boldsymbol{y}_w|} \log \pi_\theta(\boldsymbol{y}_w \mid \boldsymbol{x})\\
&-
\frac{\beta}{|\boldsymbol{y}_l|} \log \pi_\theta(\boldsymbol{y}_l \mid \boldsymbol{x})
-
\bigl(\gamma + \alpha M^\ast(x,\boldsymbol{y}_w,\boldsymbol{y}_l)\bigr)
\Bigr),
\end{aligned}
\end{equation}

% \begin{equation}
% \begin{aligned}
% \mathcal{L}_{\mathrm{alphaDPO}}(\theta;\mathcal{D})
% &=
% -\log\sigma\!\Bigl(
% \beta\!\left[
% \log\tfrac{\pi_\theta(y_w\mid x)}{\pi_\theta(y_l\mid x)}
% \right]\\
% &-
% \bigl[\gamma + \alpha M^\ast(x,y_w,y_l)\bigr]
% \Bigr),
% \end{aligned}
% \end{equation}
where $\gamma + \alpha M^\ast(x,y_w,y_l)$ denotes an adaptive reward margin that that extends the fixed SimPO margin and adjusts the required separation between preferred and dispreferred responses based on instance-specific discrepancies.

%discrepancy term. This adaptive margin allows AlphaDPO to better capture heterogeneous preference strength across samples and consistently improves alignment quality over DPO and SimPO.

\section{Experimental Details} \label{app:experimental_details}

\paragraph{Efficient Algorithm for SPO.} To efficiently solve the SPO optimization problem in Eq.~\eqref{eq:penalized_opt}, we develop an efficient algorithm following previous work~\citep{shen2024seal}. Since LLMs have a large number of parameters, solving the lower-level problem can be computationally expensive. However, due to their vast parameter space, these models can effectively fit the SFT dataset. Therefore, we can reasonably assume that the loss of the lower-level problem is a relatively small and reasonable value, denoted as \(\hat{L}_{\text{sft}}^*\). In our experiments, we set \(\hat{L}_{\text{sft}}^*\) as the minimum SFT loss obtained after fine-tuning the base model on the SFT dataset for one epoch.

\begin{algorithm}[t]
\caption{Efficient Algorithm for SPO}\label{alg:eff_bilevel_dpo}
\begin{algorithmic}[1]
\STATE \textbf{Input}: Preference dataset \(\mathcal{D}_{\mathrm{tr}}\), SFT dataset \(\mathcal{D}_{\mathrm{sft}}\), initial parameters \(\boldsymbol{\theta}_0\), penalty coefficient \(\lambda\), \(\eta_{\theta}\).
\STATE Initialize \(\boldsymbol{\theta}_0 \leftarrow \boldsymbol{\theta}_0\).
\FOR{\(t = 0, 1, \dots, T-1\)}
    % \STATE Initialize \(\boldsymbol{\theta}^{\prime(0)}_t \leftarrow \boldsymbol{\theta}_t\).
    \STATE Set \(\mathcal{L}_{\mathrm{sft}}^{*}(\boldsymbol{x}, \boldsymbol{y}_w) \approx \hat{L}_{\text{sft}}^*\).
    \STATE Sample mini-batch \((\boldsymbol{x}, \boldsymbol{y}_w, \boldsymbol{y}_l)\).
    \STATE Update \(\boldsymbol{\theta}_{t+1}\) via Eq.~\eqref{eq:variable_updates}.
\ENDFOR
\STATE \textbf{Output}: Parameters \(\boldsymbol{\theta}_{T-1}\).
\end{algorithmic}
\end{algorithm}

% 描述Preference Dataset
\paragraph{Details of Training Datasets.} For instruction-following tasks, we utilize the SFT split for supervised fine-tuning and preference split of UltraFeedback dataset~\citep{cui2023ultrafeedback} for preference learning, respectively. For reasoning tasks, we first sample 10 responses for each question in GSM8K~\citep{cobbe2021training} from the Qwen2.5-0.5B-Instruct model and the LLaMA3.2-1B-Instruct model with a temperature of 0.7 to obtain a teacher dataset. Then we filter out the responses that the pass@k metric~\citep{lyu2025top} is lower than 30\%, and finally we obtain the SFT dataset for Qwen2.5-0.5B-Instruct and LLaMA3.2-1B-Instruct, respectively. Next, we train these model using corresponding SFT dataset to obtain the SFT checkpoint. Finally, we sample 10 responses for each question in GSM8K from the SFT checkpoint using temperature sampling with a temperature of 0.7 to construct the final preference dataset.

\paragraph{Evaluation Protocols.} For instruction-following tasks, we evaluate the models on the AlpacaEval~\citep{dubois2024length} benchmark, which contains 805 diverse prompts covering various categories. We use GPT-4o-mini as the evaluator to compare model responses with reference answers, and report win rate and length controlled win rate (LC) metrics. For reasoning tasks, we evaluate the models on the GSM8K test set~\citep{cobbe2021training} using the accuracy, where a response is considered correct if the final answer matches the ground truth.

\paragraph{Training and Inference Details.} For Qwen-2.5-0.5B and LLaMA-3.2-1B, all experiments are conducted on 2 NVIDIA 4090 GPUs using CUDA 12.6 and PyTorch 2.6.0~\citep{imambi2021pytorch}, while for LLaMA-3-8B, we use 8 NVIDIA A100 GPUs. All models are fine-tuned using the AdamW optimizer with mixed-precision training in bfloat16. During inference, we utilize vLLM~\citep{kwon2023efficient} for efficient decoding, the version used is v0.8.3.

\paragraph{Hyperparameters Setting.} In supervised learning stage, we fine-tune the models using the UltraFeedback dataset~\citep{cui2023ultrafeedback} with a batch size of 128 and a maximum sequence length of 1024 for both the SFT and preference learning stages. In both stages, we use a cosine learning rate schedule with 5\% warmup steps, training for 1 epoch. The learning rate is set to $2 \times 10^{-5}$ during the SFT stage for both instruction-following and reasoning tasks.

During preference learning stage, we found that setting \(\beta\) between 0.1 and 0.5 generally yields good performance across various tasks and models except for SimPO paper~\citep{meng2024simpo} and AlphaDPO~\citep{wu2025alphadpo}, which requires a larger \(\beta\) value to achieve optimal results, likely due to its length-normalized reward structure. Also, this observation is reported in SimPO~\citep{meng2024simpo} and AlphaDPO~\citep{wu2025alphadpo}. The batch size is 128, and we train for 1 epoch across all methods. Since the goal of our experiments is to evaluate the effectiveness of SPO in improving preference optimization, we also carefully tune the method-specific hyperparameters of all baseline methods. For SPO, since different methods may have different scales of the reward function, we tune the hyperparameter \(\gamma\) and \(\lambda\) for each SPO variant. All the hyperparameter search ranges used in our experiments is summarized in Table~\ref{tab:hyperparameters}.

\paragraph{License and Terms of Use.} We utilize open-source artifacts in this work. Specifically, the UltraFeedback dataset is licensed under the MIT License. The Qwen-2.5 models are under the Apache 2.0 License, and the LLaMA-3/3.2 models are under the LLAMA 3/3.2 Community License. The GSM8K dataset is released under the MIT License. All these artifacts are used in accordance with their respective terms for research purposes.

\begin{table*}[ht]
\vspace{-1em}
\caption{Hyperparameters search range for different preference optimization methods}
\label{tab:hyperparameters}
\centering
\resizebox{\textwidth}{!}{
\small
\begin{tabular}{lll}
\toprule
\textbf{Method} & \textbf{Method-specific Hyperparameter} & \textbf{SPO Hyperparameter} \\ \midrule
\multirow{2}{*}{DPO} & \multirow{2}{*}{$\beta \in [0.01, 0.05, 0.1]$} & $\gamma \in [0.1, 1.0]$ \\
 & & $\lambda \in [0.0, 0.001, 0.01, 0.1, 1.0]$ \\ \midrule

\multirow{2}{*}{CPO} & \multirow{2}{*}{$\beta \in [0.01, 0.05, 0.1]$, $\lambda = 1.0$} & $\gamma \in [0.1, 1.0]$ \\
 & & $\lambda \in [0.0, 0.001, 0.01, 0.1, 1.0]$ \\ \midrule

\multirow{2}{*}{RDPO} & \multirow{2}{*}{$\beta \in [0.01, 0.05, 0.1]$, $\alpha \in [0.01, 0.02, 0.05]$} & $\gamma \in [0.1, 1.0]$ \\
 & & $\lambda \in [0.0, 0.001, 0.01, 0.1, 1.0]$ \\ \midrule

\multirow{2}{*}{DPOP} & \multirow{2}{*}{$\beta \in [0.01, 0.05, 0.1]$, $\lambda \in [1.0, 10.0, 50.0]$} & $\gamma \in [0.1, 1.0]$ \\
 & & $\lambda \in [0.0, 0.001, 0.01, 0.1, 1.0]$ \\ \midrule

\multirow{2}{*}{SimPO} & \multirow{2}{*}{$\beta \in [1.0, 2.0, 3.0]$, $\gamma \in [0.1, 1.0, 2.0]$} & $\gamma \in [0.1, 1.0]$ \\
 & & $\lambda \in [0.0, 0.001, 0.01, 0.1, 1.0]$ \\ \midrule

\multirow{2}{*}{AlphaDPO} & \multirow{2}{*}{$\beta \in [1.0, 2.0, 3.0]$, $\alpha \in [0.05, 0.1, 0.2]$, $\gamma \in [0.1, 0.2, 0.5]$} & $\gamma \in [0.1, 1.0]$ \\
 & & $\lambda \in [0.0, 0.001, 0.01, 0.1, 1.0]$ \\

\bottomrule
\end{tabular}
}
\end{table*}

\section{Additional Experiments} \label{app:additional_experiments}

\subsection{Empirical Investigation of Theorem~\ref{thm:prob_change}} \label{Empirical Investigation of Theorem 1}

% Inserting figures to visualize the dynamics using subfig package
\begin{figure*}[ht]
\centering
\subfloat[\label{fig:log_probs}]{\includegraphics[width=0.33\textwidth]{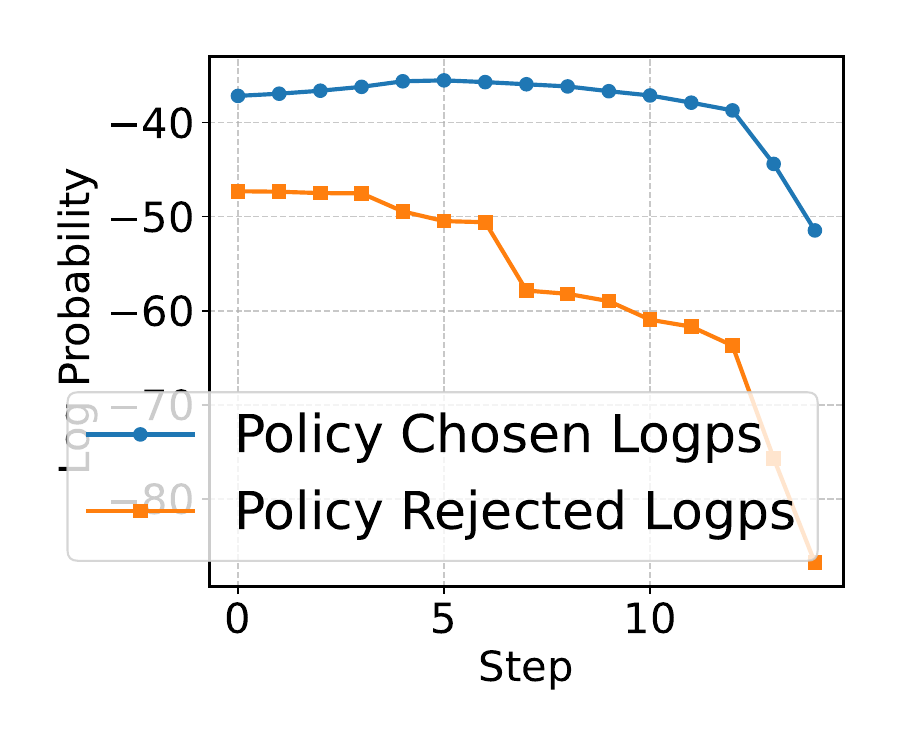}}\hfill
\subfloat[\label{fig:grad_terms}]{\includegraphics[width=0.33\textwidth]{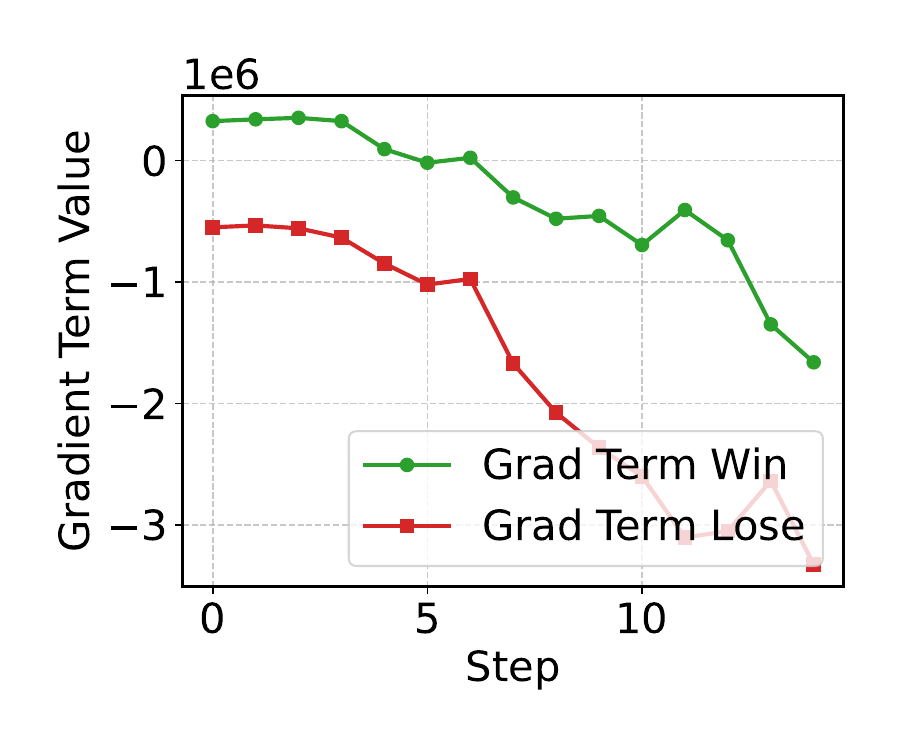}}\hfill
\subfloat[\label{fig:log_delta_ratio}]{\includegraphics[width=0.33\textwidth]{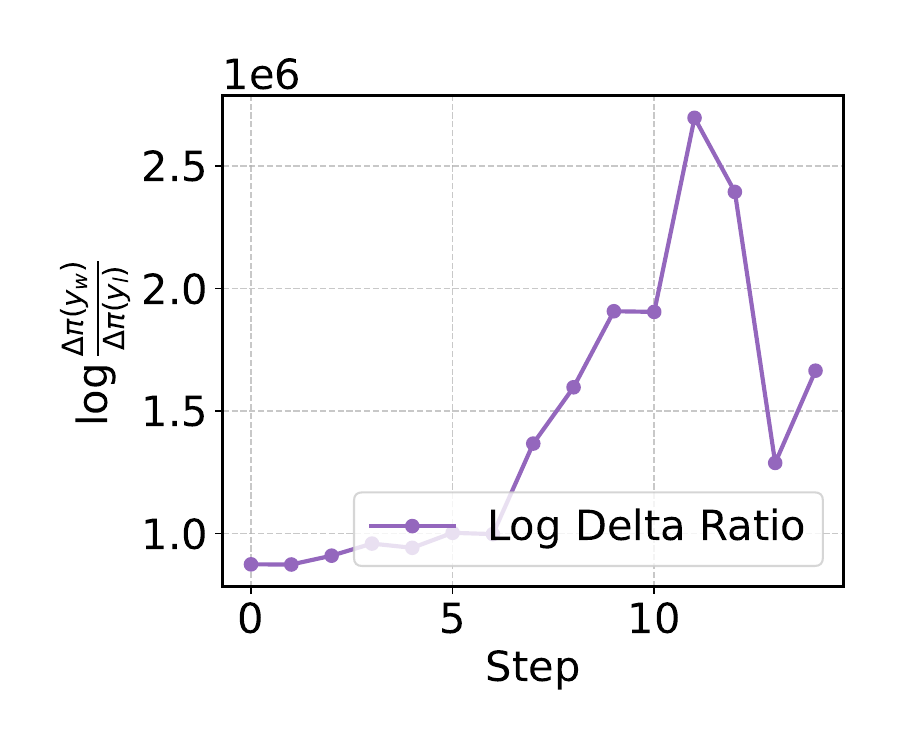}}
\caption{Visualization of DPO dynamics: (a) Log probabilities of preferred (blue) and dispreferred (orange) samples over steps, showing the divergence in probabilities; (b) Gradient terms for preferred (green) and dispreferred (red) samples in Eqs.~\eqref{eq:delta_pi_y_w} and ~\eqref{eq:delta_pi_y_l}, illustrating their optimization trends; (c) The value of \(\log \frac{\Delta \pi_{\boldsymbol{\theta}}(\boldsymbol{y}_w|\boldsymbol{x})}{\Delta \pi_{\boldsymbol{\theta}}(\boldsymbol{y}_l|\boldsymbol{x})}\) (purple) over steps, representing the relative probability changes between chosen and rejected samples.}
\label{fig:dpo_dynamics}
\vspace{-4pt}
\end{figure*}

\begin{figure*}[ht]
\centering
\subfloat[\label{fig:prob_changes}]{\includegraphics[width=0.32\textwidth]{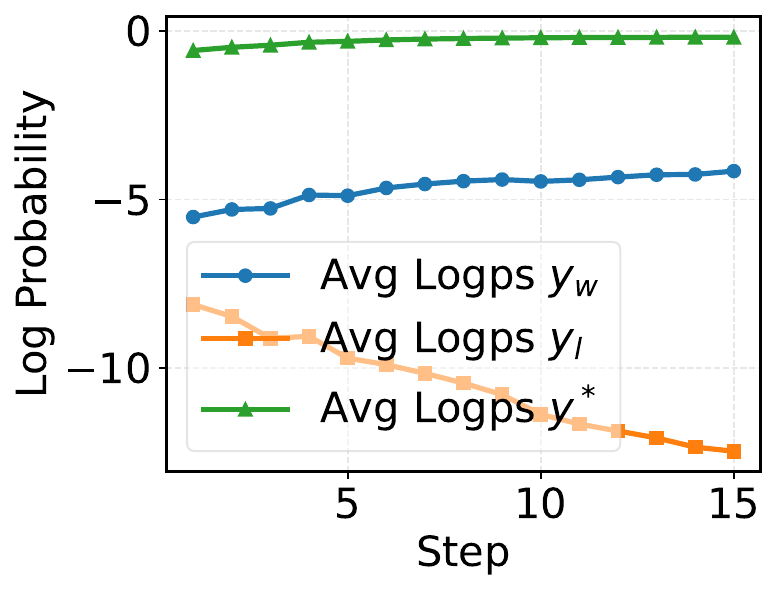}}\hspace{0.01\textwidth}
\subfloat[\label{fig:prob_changes_ours}]{\includegraphics[width=0.32\textwidth]{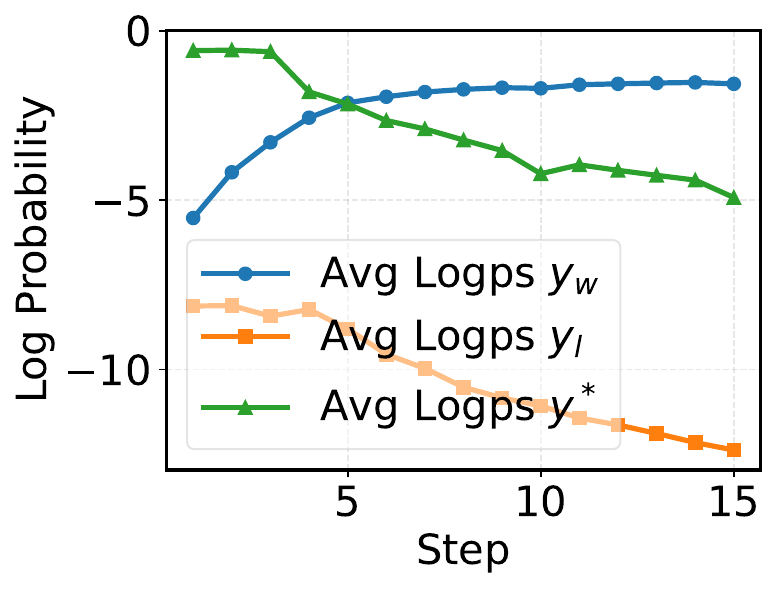}}
\caption{Visualization of Probability Mass Redistribution: (a) Average log-probability changes of \( \mathcal{D}_w \), \( \mathcal{D}_l \), and \( \boldsymbol{y}^* \) over training steps under vanilla DPO; (b) Comparison of log-probability changes under SPO.}
\label{fig:thm2}
\end{figure*}

To understand Theorem~\ref{thm:prob_change}, we visualize the optimization behavior of standard DPO using a probing dataset with Qwen-2.5-0.5B-Instruct model in Figure~\ref{fig:dpo_dynamics}. Figure~\ref{fig:log_probs} plots the log-probability of the preferred (blue) and dispreferred (orange) samples over steps. It shows that the preferred sample \( \boldsymbol{y}_w \) initially exhibits a slow increase in log-probability before declining, while the dispreferred sample \( \boldsymbol{y}_l \) consistently decreases. This indicates a complex probability dynamic where both samples ultimately experience a decline, which is a counter-intuitive behavior in preference learning. To understand this phenomenon, Figure~\ref{fig:grad_terms} illustrates the corresponding gradient terms for the preferred (green) and dispreferred (red) samples as in Theorem~\ref{thm:prob_change}, defined as \( \boldsymbol{g}_w^\top (\boldsymbol{g}_w - \boldsymbol{g}_l) \) and \( \boldsymbol{g}_l^\top (\boldsymbol{g}_w - \boldsymbol{g}_l) \), respectively. The gradient term for the preferred sample starts positive before turning negative, which align with the dynamics of \( \pi_{\boldsymbol{\theta}}(\boldsymbol{y}_w|\boldsymbol{x}) \), while the gradient term for the dispreferred sample remains negative throughout the steps. These trends further validate the correctness of Theorem~\ref{thm:prob_change}, where the sign of the gradient terms govern the direction of probability changes.

We also visualize the log ratio \(\log \frac{\Delta \pi_{\boldsymbol{\theta}}(\boldsymbol{y}_w|\boldsymbol{x})}{\Delta \pi_{\boldsymbol{\theta}}(\boldsymbol{y}_l|\boldsymbol{x})}\) in Figure~\ref{fig:log_delta_ratio} to illustrate the relative probability changes between the preferred and dispreferred samples. The ratio remains positive over steps, which indicates that the probability change for the preferred sample \( \boldsymbol{y}_w \) consistently exceeds that of the dispreferred sample \( \boldsymbol{y}_l \), even when both probabilities decline as observed in Figure~\ref{fig:log_probs}. This trend aligns with Corollary~\ref{corollary:relative_change}, which guarantees \(\Delta \pi_{\boldsymbol{\theta}}(\boldsymbol{y}_w|\boldsymbol{x}) - \Delta \pi_{\boldsymbol{\theta}}(\boldsymbol{y}_l|\boldsymbol{x}) \geq 0\).

\subsection{Empirical Investigation of Theorem~\ref{thm:mass_shift}} \label{Empirical Investigation of Theorem 2}
Since the neural tangent kernel may change during training, making it challenging to directly observe the similarity between gradients as stated in Theorem~\ref{thm:mass_shift}, we design a simple experiment to empirically validate the theorem using Vanilla DPO. We use the query ``\( 11 \times 12 = ? \)" (correct answer: 132). We sample 500 responses from Qwen2.5-0.5B-Instruct. Then the highest probability response \(\boldsymbol{y}_{*}\) in all sampled responses is treated as \( \mathcal{D}^*=\{ \boldsymbol{y}^* \} \). Next, we categorize the sampled responses into correct and incorrect groups based on whether they match the correct answer except for \( \boldsymbol{y}_{*} \). The highest probability correct response in all correct responses is selected as \( \boldsymbol{y}_w \) to form the preferred set \( \mathcal{D}_{\mathrm{sw}} = \{ \boldsymbol{y}_w\} \), and the highest probability incorrect response is selected as \( \boldsymbol{y}_l \) to form the dispreferred set \( \mathcal{D}_{\mathrm{sl}} = \{ \boldsymbol{y}_l \} \). We then train with DPO on this dataset and monitor the average log-probability changes of \( \mathcal{D}_{\mathrm{sw}} \), \( \mathcal{D}_{\mathrm{sl}} \), and \(\mathcal{D}^*\). The results are shown in Figure~\ref{fig:thm2}.
% \begin{figure*}[ht]
% \centering
% \subfloat[\label{fig:prob_changes}]{\includegraphics[width=0.4\textwidth]{figs/training_dynamics_thm2_dpo.pdf}}\hfill
% \subfloat[\label{fig:prob_changes_ours}]{\includegraphics[width=0.4\textwidth]{figs/training_dynamics_thm2_sft.pdf}}\hfill
% \caption{Visualization of Probability Mass Redistribution: (a) Average log-probability changes of \( \mathcal{D}_w \), \( \mathcal{D}_l \), and \( y^* \) over training steps under vanilla DPO; (b) Comparison of log-probability changes under SPO.}
% \label{fig:thm2}
% \end{figure*}

Figure~\ref{fig:prob_changes} illustrates the average log-probability dynamic over training steps. The average log-probability of \( \mathcal{D}_{\mathrm{sw}} \) increases, while \( \mathcal{D}_{\mathrm{sl}} \) decreases, aligning with DPO's intended goal. However, this process also leads to an unintended increase in the probability of \( \mathcal{D}^* \), which is primarily attributed to the high initial probability of \( \mathcal{D}^* \) from the initial model. This phenomenon demonstrates that samples with high initial probabilities are more likely to absorb lost probability mass. However, if \( \mathcal{D}^* \) does not align with our intended preference optimization goal, this increase in its probability represents an undesirable side effect which may lead to unexpected behaviors. These findings support the second case of Theorem~\ref{thm:mass_shift}, where \( \Delta \pi_{\boldsymbol{\theta}}(\boldsymbol{y}_l|\boldsymbol{x}) < 0 \) (as seen in \( \mathcal{D}_{\mathrm{sl}} \)) leads to probability mass shifting to \( \mathcal{D}_{\mathrm{sw}} \) and \( \mathcal{D}^* \). While the increase in \( \mathcal{D}_{\mathrm{sw}} \)'s probability aligns with preference optimization objective, the rise in \( \mathcal{D}^* \)'s probability highlights a catastrophic preference shift. Thus, it is important add additional constraint during preference optimization.

Figure~\ref{fig:prob_changes_ours} presents the log-probability dynamics under our proposed SPO method. Compared to Vanilla DPO, DPO+SPO effectively suppresses the unintended increase in \( \mathcal{D}^* \)'s probability while still promoting the desired increase in \( \mathcal{D}_{\mathrm{sw}} \)'s probability and decrease in \( \mathcal{D}_{\mathrm{sl}} \)'s probability. This indicates that SPO successfully mitigates the failure mode observed in Vanilla DPO, aligning with our theoretical analysis.

\subsection{Visualization of training dynamics}\label{app:visualization_of_training_dynamics}

\begin{figure*}[t]
\centering
% \subfloat[\label{fig:dpo_training_dynamics}]{\includegraphics[width=0.99\textwidth]{figs/dpo_spo_training_dynamics.pdf}}\hfill % hfill表示水平填充
\subfloat[Training dynamics of DPOP\label{fig:dpop_training_dynamics}]{\includegraphics[width=0.99\textwidth]{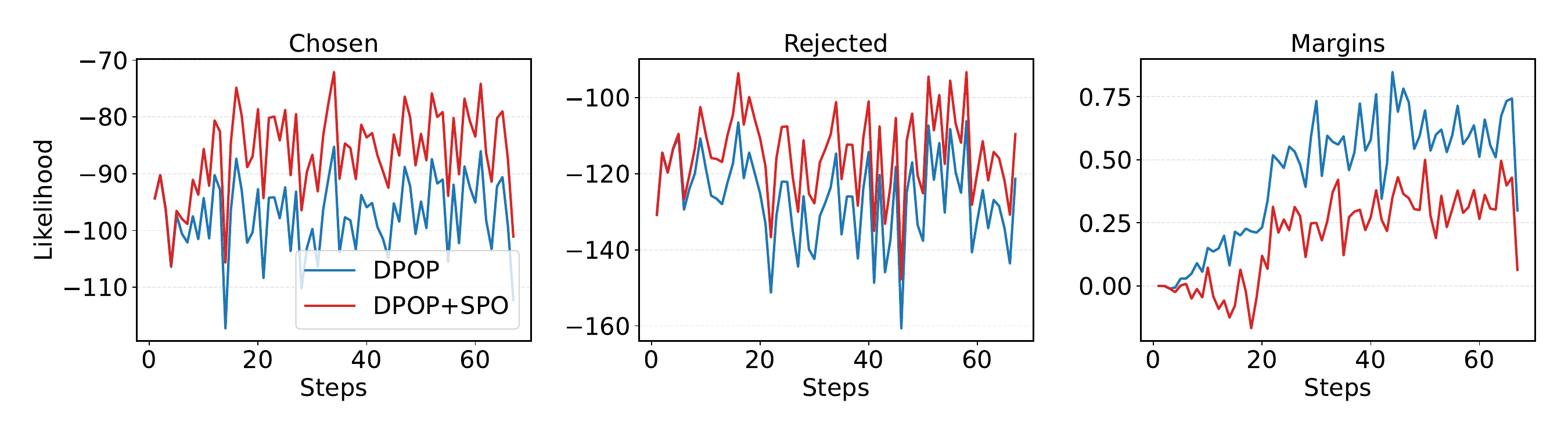}}\hfill
\subfloat[Training dynamics of RDPO\label{fig:rdpo_training_dynamics}]{\includegraphics[width=0.99\textwidth]{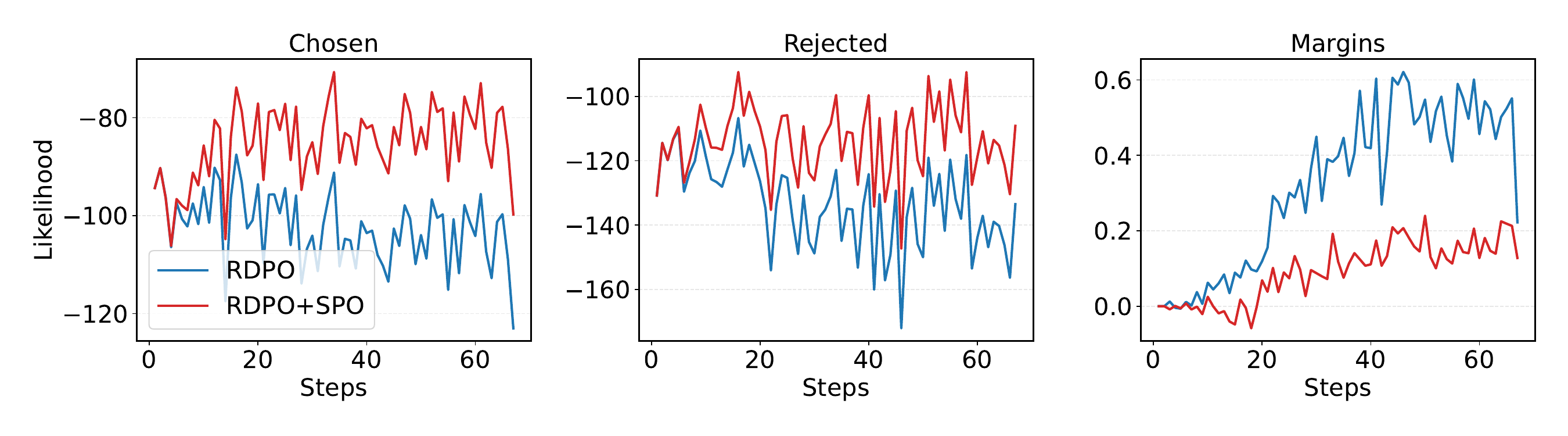}}\hfill
\subfloat[Training dynamics of SimPO\label{fig:simpo_dpo_training_dynamics}]{\includegraphics[width=0.99\textwidth]{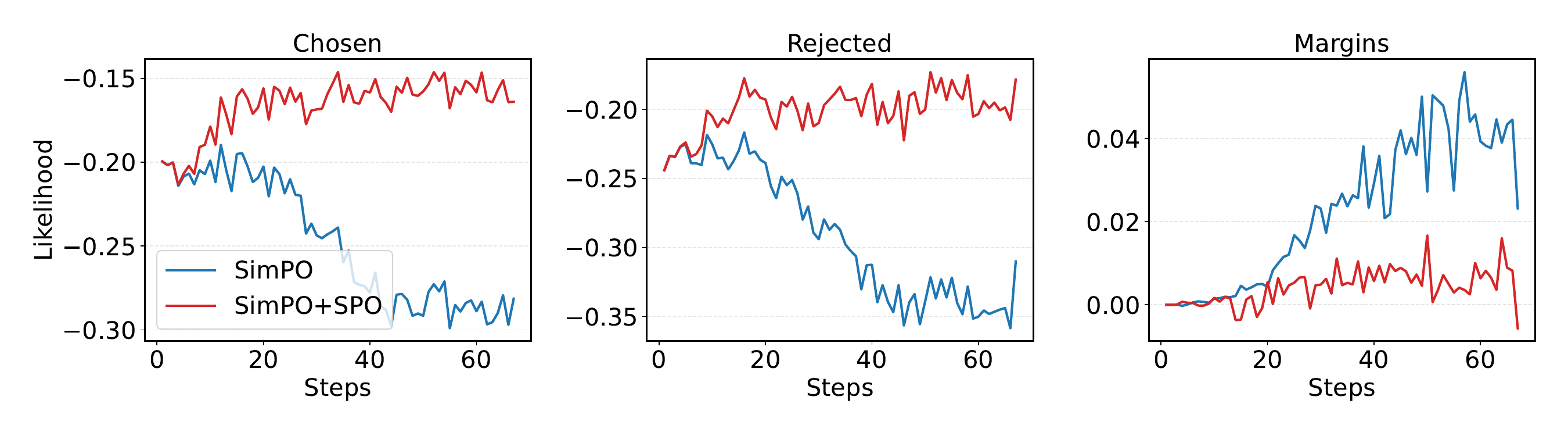}}\hfill
\subfloat[Training dynamics of Alpha DPO\label{fig:alpha_dpo_training_dynamics}]{\includegraphics[width=0.99\textwidth]{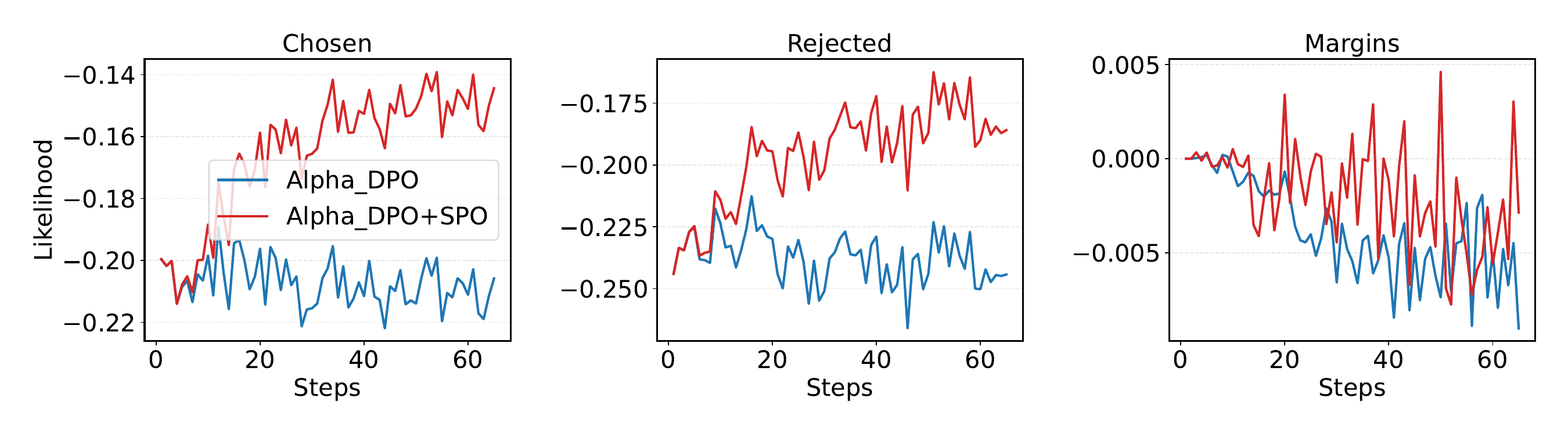}}\hfill
\caption{Visualization of Training Dynamics on GSM8K.}
\label{fig:training_dynamics_all}
\end{figure*}

We further visualize the training dynamics of DPO, CPO,  RDPO, DPOP and AlphaDPO on GSM8K with Qwen2.5-0.5B-Instruct in Figure~\ref{fig:training_dynamics_all}. We can observe similar trends that almost all of these methods suffer from the decline of preferred sample probability during training, while the SPO-enhanced versions effectively mitigate this issue and improve the reasoning accuracy according to Table~\ref{tab:main_results}. Notably, most of these methods tend to increase the reward margin during training, but it does not lead to a good performance. Instead, SPO-enhanced version maintains a relative reasonable reward margin, which is attributed to the bilevel structure of SPO. This indicates that the SPO framework can effectively mitigate the issue of probability decline for preferred samples while maintaining a reasonable reward margin, which is crucial for preference optimization. 

Among these methods, we can also see that the DPOP method exhibits less severe probability decline for preferred samples compared to other methods, this is because DPOP introduces an additional penalty term to prevent the preferred response from falling below its reference-model likelihood, which helps to maintain generative stability to some extent. However, DPOP fundamentally differs from our SPO framework in both its theoretical motivation and optimization mechanism. Specifically, DPOP acts as a local heuristic penalty that prevents individual response probabilities from dropping significantly below their reference model. While this reduces likelihood displacement, it does not fundamentally address the conflict between discriminative and generative objectives. In contrast, SPO dynamically constrains the preference optimization trajectory within the safe aligment region $\mathfrak{R}$, which address the limitation of existing BT-style methods. Therefore, according to our experiments in Table~\ref{tab:main_results} and the training dynamics in Figure~\ref{fig:dpop_training_dynamics}, we can see DPOP still benefits from the SPO framework to further enhance its performance during training.

\subsection{Extent to SFT-free preference optimization}\label{app:extend_to_sft_free}
Existing preference optimization piplines first warmup the base model on SFT dataset to obtain a good initialization, and then apply preference learning on SFT checkpoint. However, this may be computationally expensive, and it is desirable to directly perform preference optimization on the base model. To investigate whether SPO can still be effective in SFT-free setting, we extend various preference optimization methods with SPO to directly fine-tune base models. We conduct experiments and evaluate on AlpacaEval and GSM8K using Qwen2.5-0.5B model and LLaMA3.2-1B. we compare these methods width strong baseline ORPO~\citep{hong2024orpo}, which steers models towards human preferences without supervised fine-tuning. The results are summarized in Table~\ref{tab:sft_free_results}. From the results, we can observe that SPO can effectively extend existing BT-style preference optimization methods to SFT-free setting, demonstrating its broad potential. While some methods like DPO and RDPO do not surpass the specialized SFT-free baseline ORPO, we surprisingly find that SimPO+SPO and AlphaDPO+SPO significantly outperform ORPO by a large margin, demonstrating the strong capability of SPO in stabilizing preference optimization even without SFT initialization.

\begin{table}[ht]
\centering
\caption{Performance comparison of different preference optimization methods in SFT-free setting. The best and second-best results are highlighted in \textbf{bold} and \underline{underline}, respectively.}
\vspace{-6pt}
\label{tab:sft_free_results}
\resizebox{0.48\textwidth}{!}{
\begin{tabular}{l ccc}
\toprule
\multirow{3}{*}{\textbf{Method}}
& \multicolumn{3}{c}{\textbf{Qwen-2.5-0.5B}} \\
\cmidrule(lr){2-4}

& \multicolumn{2}{c}{\textbf{AlpacaEval 2}} & \textbf{GSM8K} \\
\cmidrule(lr){2-3}\cmidrule(lr){4-4}

& \textbf{WR (\%)} & \textbf{LC (\%)} & \textbf{Acc. (\%)} \\
\midrule

ORPO
& 4.4 & 4.3 & 37.0 \\
\midrule

DPO + SPO
& 2.8 & 2.8 & 33.6 \\
CPO + SPO
& 2.5 & 2.4 & 43.2 \\
DPOP + SPO
& 2.8 & 2.2 & 34.5 \\
RDPO + SPO
& 2.2 & 2.1 & 34.1 \\
SimPO + SPO
& \underline{5.2} & \underline{4.8} & \underline{51.3} \\
AlphaDPO + SPO
& \textbf{5.3} & \textbf{5.2} & \textbf{52.5} \\
\bottomrule
\end{tabular}
}
\vspace{-6pt}
\end{table}

\subsection{Evaluation on downstream tasks}\label{app:evaluation_on_downstream_tasks}

To further validate the effectiveness of SPO, we also evaluate the models fine-tuned with different preference optimization methods on downstream tasks, including ARC~\citep{clark2018think}, BoolQ~\citep{clark2019boolq}, PIQA~\citep{bisk2020piqa}, TruthfulQA~\citep{lin2022truthfulqa}, HellaSwag~\citep{zellers2019hellaswag}, MUSR~\citep{zhong2023musr}, IFEval~\citep{zhou2023instructionfollowingevaluationlargelanguage}. These tasks cover a wide range of language model capabilities, including scientific reasoning, commonsense reasoning, and instruction following. We evaluate all models using zero-shot prompting and the results are summarized in Table~\ref{tab:results_on_downstream_tasks} and Table~\ref{tab:combined_avg_rank}. From the average rank results in Table~\ref{tab:combined_avg_rank}, we can clearly see the superiority of SPO-enhanced methods over their vanilla methods. Specifically, the recent 2 state-of-the-art methods AlphaDPO and SimPO achieve significant performance improvements when combined with SPO, ranking the top two positions. These results further validate the effectiveness and generalizability of SPO.

\begin{table}[ht]
\centering
\caption{The average rank of different methods across Qwen0.5B, Llama1B and Llama8B models. The best and second-best results are highlighted in \textbf{bold} and \underline{underline}, respectively.}
\label{tab:combined_avg_rank}
\resizebox{0.45\textwidth}{!}{
\begin{tabular}{lcccc}
\toprule
\textbf{Model} & \textbf{0.5B} & \textbf{1B} & \textbf{8B}  & \textbf{Avg.} \\
\midrule
AlphaDPO+SPO & \textbf{4.9} & \textbf{5.4} & \underline{5.3} & \textbf{5.2} \\
SimPO+SPO & 6.4 & \underline{6.3} & \textbf{4.6} & \underline{5.8} \\
DPOP+SPO & 6.3 & 7.3 & 6.3 & 6.6 \\
DPO+SPO & 6.7 & 7.9 & 5.9 & 6.8 \\
SimPO & 7.0 & 6.7 & 6.9 & 6.9 \\
AlphaDPO & \underline{5.3} & 6.4 & 9.0 & 6.9 \\
RDPO+SPO & 7.0 & 6.1 & 7.6 & 6.9 \\
DPO & 7.7 & 6.0 & 7.1 & 7.0 \\
CPO+SPO & 7.9 & 7.0 & 6.4 & 7.1 \\
RDPO & 7.4 & 6.9 & 7.6 & 7.3 \\
CPO & 8.4 & 7.1 & 6.9 & 7.5 \\
DPOP & 7.4 & 8.4 & 6.7 & 7.5 \\
SFT & 8.6 & 9.4 & 10.9 & 9.6 \\
\bottomrule
\end{tabular}
}
\end{table}

\subsection{Case Study}\label{app:case_study}
To illustrate the \textit{preference shift} phenomenon, we present failure case studies after applying different BT-style preference optimization methods without SPO. Since reasoning tasks are particularly sensitive to this shift, we select several cases from GSM8K using LLaMA3-8B models. Specifically, we focus on SimPO and AlphaDPO because these two methods exhibit the most significant performance degradation in our main results.

As shown in Table~\ref{tab:detailed_case_study_1}, both models initially demonstrate correct reasoning steps but ultimately devolve into infinite loops of repetitive tokens. This behavior indicates a clear preference shift towards these degenerate modes, such as repetitive tokens (e.g., \texttt{\textbackslash boxed\{...\}}) in SimPO and repeated phrases (e.g., \texttt{Wait, maybe I'm misunderstanding the problem.}) in AlphaDPO. These case studies highlight the critical need for methods like SPO to maintain alignment with human preferences during preference optimization. 

However, when we integrate SPO into these methods, such failure cases are significantly mitigated, leading to more accurate responses. As shown in Table~\ref{tab:detailed_case_study_2}, both SimPO+SPO and AlphaDPO+SPO generate correct and coherent reasoning steps that lead to the right answer without falling into repetitive loops. This demonstrates SPO stabilizes preference optimization and prevent catastrophic preference shift. 

Moreover, we analyze the responses of AlphaDPO and SimPO-fine-tuned LLaMA3-8B models on GSM8K and find that there are a large portion of responses exhibiting repetitive patterns, leading to severe performance degradation.  According to the High Inflow Problem discussed in~\citet{fu2021a}, when probability mass is removed from both preferred and dispreferred responses, the redistributed mass does not reliably flow toward correct solutions but instead concentrates on a small set of high-inflow tokens that are reachable from many contexts with high transition probability. This uncontrolled concentration induces self-reinforcing transition loops, leading to repetitive high-probability responses and ultimately causing a catastrophic collapse in model behavior. 
So according to Theorem~\ref{thm:prob_change} and Theorem~\ref{thm:mass_shift}, the probability mass instead flow to these high initial probability samples with undesirable repetitive patterns, resulting in the observed failure cases.

%This further validates the effectiveness of SPO in mitigating the catastrophic preference shift and enhancing the overall performance of preference optimization methods.

\begin{table*}[ht] 
\centering 
\caption{Downstream tasks evaluation of different preference optimization methods.} 
\label{tab:results_on_downstream_tasks}
\resizebox{0.99\textwidth}{!}{
\begin{tabular}{l c c c c c c c c c}
\toprule
% \textbf{Method} & \textbf{arc\_easy} & \textbf{boolq} & \textbf{piqa} & \textbf{truthfulqa} & \textbf{hellaswag} & \textbf{rte} & \textbf{mmlu\_pro} & \textbf{musr} & \textbf{Avg. Rank} \\

\textbf{Method} & \textbf{ARC\_easy} & \textbf{BoolQ} & \textbf{PIQA} & \textbf{TruthfulQA} & \textbf{HellaSwag} & \textbf{MUSR} & \textbf{IFEval} & \textbf{Avg. Rank} \\

\midrule
\multicolumn{9}{c}{\textbf{Qwen2.5-0.5B}} \\
\hline
SFT & 56.3 & 57.3 & 70.5 & 35.4 & 52.4 & 34.4 & 14.2 & 8.57 \\
DPO & 56.2 & 60.6 & 70.2 & 34.7 & 53.6 & 34.7 & 12.1 & 7.71 \\
DPO + SPO & 59.2 & 60.6 & 70.2 & 35.6 & 54.2 & 34.7 & 10.7 & 6.71 \\
CPO & 56.4 & 61.4 & 69.9 & 37.2 & 53.2 & 34.5 & 11.5 & 8.43 \\
CPO + SPO & 57.3 & 61.9 & 70.0 & 36.8 & 53.5 & 34.3 & 11.6 & 7.86 \\
DPOP & 58.7 & 59.6 & 70.2 & 35.8 & 53.2 & 34.7 & 12.9 & 7.43 \\
DPOP + SPO & 58.5 & 60.3 & 70.3 & 35.6 & 53.6 & 34.9 & 11.8 & 6.29 \\
RDPO & 62.6 & 61.0 & 70.0 & 34.9 & 53.9 & 34.7 & 11.6 & 7.43 \\
RDPO + SPO & 58.0 & 59.9 & 70.1 & 35.5 & 54.2 & 34.9 & 12.9 & 7.00 \\
SimPO & 61.4 & 45.4 & 69.8 & 37.6 & 55.7 & 32.0 & 17.6 & 7.00 \\
SimPO + SPO & 62.2 & 50.6 & 70.5 & 35.4 & 55.9 & 32.8 & 16.6 & 6.43 \\
AlphaDPO & 60.7 & 50.7 & 70.5 & 37.5 & 56.1 & 32.5 & 17.0 & 5.29 \\
AlphaDPO + SPO & 64.0 & 60.1 & 70.0 & 38.2 & 56.4 & 31.2 & 17.2 & 4.86 \\

\midrule
\multicolumn{9}{c}{\textbf{LLaMA3.2-1B}} \\
\hline
SFT & 62.0 & 59.3 & 74.8 & 36.2 & 64.9 & 35.8 & 12.2 & 9.43 \\
DPO & 61.6 & 60.4 & 74.8 & 39.0 & 68.2 & 37.0 & 14.8 & 6.00 \\
DPO + SPO & 62.0 & 57.1 & 74.8 & 37.0 & 66.7 & 37.2 & 13.9 & 7.86 \\
CPO & 60.2 & 60.7 & 74.4 & 37.0 & 66.8 & 37.4 & 15.2 & 7.14 \\
CPO + SPO & 61.8 & 65.2 & 74.3 & 36.9 & 66.4 & 36.0 & 17.4 & 7.00 \\
DPOP & 60.1 & 51.9 & 74.8 & 38.6 & 67.6 & 37.7 & 12.8 & 8.43 \\
DPOP + SPO & 62.3 & 58.5 & 74.9 & 36.5 & 66.0 & 38.5 & 13.9 & 7.29 \\
RDPO & 61.3 & 59.3 & 75.1 & 39.0 & 68.2 & 35.7 & 14.6 & 6.86 \\
RDPO + SPO & 62.8 & 61.4 & 74.9 & 36.8 & 66.4 & 37.8 & 14.4 & 6.14 \\
SimPO & 63.7 & 63.2 & 72.3 & 39.9 & 70.5 & 34.3 & 12.8 & 6.71 \\
SimPO + SPO & 64.0 & 63.9 & 72.9 & 39.4 & 70.8 & 35.1 & 11.8 & 6.29 \\
AlphaDPO & 63.8 & 64.1 & 73.2 & 41.6 & 70.4 & 31.7 & 8.7 & 6.43 \\
AlphaDPO + SPO & 63.4 & 61.8 & 74.6 & 40.3 & 69.5 & 36.4 & 13.3 & 5.43 \\
\midrule
\multicolumn{9}{c}{\textbf{LLaMA3-8B}} \\
\hline
SFT & 78.9 & 76.4 & 80.7 & 35.8 & 78.8 & 38.4 & 17.4 & 10.86 \\
DPO & 80.5 & 80.9 & 80.4 & 40.3 & 80.7 & 40.9 & 20.9 & 7.14 \\
DPO + SPO & 82.1 & 83.5 & 81.7 & 39.4 & 80.1 & 41.8 & 17.2 & 5.86 \\
CPO & 78.7 & 82.9 & 81.7 & 42.1 & 80.9 & 37.3 & 17.4 & 6.86 \\
CPO + SPO & 79.0 & 81.6 & 82.0 & 41.2 & 80.9 & 39.0 & 18.7 & 6.43 \\
DPOP & 80.8 & 83.0 & 81.4 & 41.9 & 80.8 & 39.2 & 16.6 & 6.71 \\
DPOP + SPO & 80.6 & 83.8 & 81.7 & 40.1 & 79.6 & 40.7 & 17.4 & 6.29 \\
RDPO & 77.7 & 83.5 & 81.1 & 40.8 & 80.9 & 39.0 & 20.1 & 7.57 \\
RDPO + SPO & 81.9 & 82.4 & 81.5 & 39.5 & 80.2 & 40.1 & 16.3 & 7.57 \\
SimPO & 78.2 & 84.8 & 78.4 & 51.3 & 75.7 & 40.6 & 22.4 & 6.86 \\
SimPO + SPO & 79.8 & 85.3 & 80.6 & 45.6 & 82.7 & 39.9 & 23.8 & 4.57 \\
AlphaDPO & 77.3 & 79.3 & 79.9 & 45.4 & 79.6 & 37.7 & 24.0 & 9.00 \\
AlphaDPO + SPO & 79.8 & 82.9 & 81.4 & 51.2 & 83.8 & 36.2 & 24.0 & 5.29 \\
\bottomrule
\end{tabular}
}
\end{table*}

\begin{table*}[ht]
\centering
\small
\begin{tabular}{c|p{0.78\textwidth}}
\toprule
\multirow{2}{*}{\textbf{Question}} & \multicolumn{1}{p{0.78\textwidth}}{Farmer Brown has 20 animals on his farm, all either chickens or cows. They have a total of 70 legs, all together. How many of the animals are chickens?} \\
\midrule
\textbf{Method} & \textbf{Model Response} \\
\midrule

\multirow{25}{*}{\textbf{SimPO}} &
\begin{minipage}[t]{0.78\textwidth}
\vspace{1mm}
\ttfamily
\textcolor{ForestGreen}{<think>\\
First, I need to determine the total number of legs contributed by the chickens and cows separately. \\\\
Chickens typically have two legs each, and cows have four legs each. \\\\
Let's denote the number of chickens as C and the number of cows as D. \\\\
The total number of legs for chickens is 2C, and for cows, it's 4D. \\\\
The overall total number of legs is given as 70: \\
2C + 4D = 70 \\\\
Since the total number of animals is: \\
C + D = 20} \\\\
\textcolor{WildStrawberry}{Expanding: \\
2C + 4(C + D - 20) = 70 \\\\
Combining like terms: \\
2D + 22 = 20 \\
2D = -2 \\\\
Dividing both sides by -2:} \\\\
\textcolor{Red}{D = \textbackslash{}boxed\{\textbackslash{}boxed\{\textbackslash{}boxed\{...\} \\
\textbackslash{}boxed\textbackslash{}boxed\textbackslash{}boxed\textbackslash{}boxed\textbackslash{}boxed\textbackslash{}boxed\textbackslash{}boxed\textbackslash{}boxed\textbackslash{}boxed...(repeats)}
\vspace{1mm}
\end{minipage} \\

\midrule

\multirow{27}{*}{\textbf{AlphaDPO}} &
\begin{minipage}[t]{0.78\textwidth}
\vspace{1mm}
\ttfamily
\textcolor{ForestGreen}{<think>\\
Okay, so I have this math problem here. \\\\
Farmer Brown has 20 animals, all either chickens or cows. \\
Together, they have 70 legs. \\\\
Let C be the number of chickens and K the number of cows. \\\\
2C + 4K = 70} \\\\
\textcolor{WildStrawberry}{Trying substitution: \\
C = 70 - 2K \\\\
Substituting back: \\
2C + 4(70 - C) = 70 \\
2C + 280 - 4C = 70 \\
-2C = -210 \\
C = 105 \\\\
Wait, that seems like a lot of chickens. \\
That would imply negative cows.} \\\\
\textcolor{Red}{Wait, maybe I'm misunderstanding the problem. \\
Wait, maybe I'm misunderstanding the problem. \\
Wait, maybe I'm misunderstanding the problem. \\
Wait, maybe I'm misunderstanding the problem. \\
... (repeats)}
\vspace{1mm}
\end{minipage} \\

\bottomrule
\end{tabular}

\caption{Failure Cases: \textcolor{ForestGreen}{Green} indicates correct reasoning steps, \textcolor{WildStrawberry}{light red} denotes reasoning errors, and \textcolor{Red}{bold red} highlights reasoning collapse into repetitive loops, illustrating the \textit{Catastrophic Preference Shift} phenomenon.}
\label{tab:detailed_case_study_1}
\end{table*}

\begin{table*}[ht]
\centering
\small
\begin{tabular}{c|p{0.78\textwidth}}
\toprule
\multirow{2}{*}{\textbf{Question}} & \multicolumn{1}{p{0.78\textwidth}}{Farmer Brown has 20 animals on his farm, all either chickens or cows. They have a total of 70 legs, all together. How many of the animals are chickens?} \\
\midrule
\textbf{Method} & \textbf{Model Response} \\
\midrule

\multirow{35}{*}{\textbf{SimPO + SPO}} &
\begin{minipage}[t]{0.78\textwidth}
\vspace{1mm}
\ttfamily\footnotesize\setlength{\baselineskip}{0.75\baselineskip}
\textcolor{ForestGreen}{<think>\\
First, I need to determine the total number of legs contributed by the chickens and cows separately. Since there are 70 legs in total and all the animals are either chickens or cows, I can assume that each type of animal has a specific number of legs.\\\\
Chickens typically have 2 legs each, while cows have 4 legs each.\\\\
Let's denote the number of chickens as C and the number of cows as D.\\\\
The total number of legs can be expressed as: 2C + 4D = 70\\\\
Since the total number of animals is 20, I can also write: C + D = 20\\\\
Now, I'll substitute the expression for D from the second equation into the first equation: 2C + 4(20 - C) = 70\\\\
Simplifying the equation: 2C + 80 - 4C = 70\\
Which simplifies to: -2C + 80 = 70\\
Adding 2C to both sides: 2C = 10\\
Dividing both sides by 2: C = 5\\\\
Therefore, there are 5 chickens on the farm.\\
</think>\\\\
To determine how many chickens Farmer Brown has on his farm, let's break down the problem step by step.\\\\
**Given:**\\
- Total number of animals = 20; Total number of legs = 70\\
**Step 1: Express total legs.**\\
Chickens (2 legs): \( 2C \); Cows (4 legs): \( 4D \)\\
**Step 2: Set up equations.**\\
1) \( C + D = 20 \Rightarrow D = 20 - C \)\\
2) \( 2C + 4D = 70 \)\\
**Step 3: Solve.**\\
\( 2C + 4(20 - C) = 70 \Rightarrow 2C + 80 - 4C = 70 \)\\
\( -2C = -10 \Rightarrow C = 5 \)\\\\
**Conclusion:** There are \(\boxed{5}\) chickens on the farm.}
\vspace{1mm}
\end{minipage} \\

\midrule

\multirow{30}{*}{\textbf{AlphaDPO + SPO}} &
\begin{minipage}[t]{0.78\textwidth}
\vspace{1mm}
\ttfamily\footnotesize\setlength{\baselineskip}{0.75\baselineskip}
\textcolor{ForestGreen}{<think>\\
First, I need to determine the total number of legs on the farm. Since there are 70 legs in total and all the animals are either chickens or cows, I can assume that each animal has a certain number of legs.\\\\
Chickens typically have 2 legs each, while cows have 4 legs each. Let's denote the number of chickens as C and the number of cows as D.\\\\
The total number of legs can be expressed as 2C + 4D = 70. Simplifying this equation, I get 2C + 4D = 70.\\\\
To find the number of chickens, I'll subtract 4D from both sides: 2C = 70 - 4D. Finally, I'll divide both sides by 2 to solve for C: C = 35 - 2D.\\\\
Since C + D = 20, I can substitute D = 20 - C into the result: C = 35 - 2(20 - C). This gives C = 35 - 40 + 2C, which means C = 5.\\
</think>\\\\
To determine how many of Farmer Brown's animals are chickens, let's break down the problem step by step.\\\\
1. **Total Legs and Animals:**\\
- The total number of legs is **70** and total animals is **20**.\\
2. **Setting Up the Equation:**\\
- Let \( C \) represent chickens and \( D \) represent cows.\\
- \( C + D = 20 \)\\
- \( 2C + 4D = 70 \)\\
3. **Calculation:**\\
- From first equation: \( D = 20 - C \)\\
- Substitute into second: \( 2C + 4(20 - C) = 70 \)\\
- \( 2C + 80 - 4C = 70 \Rightarrow -2C = -10 \Rightarrow C = 5 \)\\\\
**Final Answer:** There are \(\boxed{5}\) chickens.}
\vspace{1mm}
\end{minipage} \\

\bottomrule
\end{tabular}

\caption{Success Cases with SPO: Both models maintain stable reasoning trajectories and provide correct answers, demonstrating that SPO effectively prevents catastrophic preference shifts.}
\label{tab:detailed_case_study_2}
\end{table*}

\end{document}